\documentclass[journal]{IEEEtran}
\usepackage{amsmath,amsfonts}
\usepackage{array}
\usepackage{textcomp}
\usepackage{stfloats}
\usepackage{url}
\usepackage{verbatim}
\usepackage{graphicx}
\usepackage{cite}
\usepackage{times}
\usepackage{graphicx}
\usepackage{amssymb}
\usepackage[normalem]{ulem}
\usepackage{multirow}
\usepackage[dvipsnames]{xcolor}
\usepackage{subfigure}
\usepackage{algorithm2e}
\usepackage{enumitem}
\usepackage{colortbl}
\usepackage{makecell}
\usepackage{arydshln}
\usepackage{xspace}
\usepackage[figuresright]{rotating}
\usepackage[pagebackref=true,breaklinks=true,colorlinks,bookmarks=false]{hyperref}

\graphicspath{{Figs/}}
\makeatletter
\DeclareRobustCommand\onedot{\futurelet\@let@token\@onedot}
\def\@onedot{\ifx\@let@token.\else.\null\fi\xspace}
\def\eg{\emph{e.g}\onedot} 
\def\ie{\emph{i.e}\onedot} 
 
\def\etc{\emph{etc}\onedot}

\def\fig{Fig\onedot} \def\figs{Figs\onedot}
\def\eq{Eq\onedot} 
\def\sec{Sec\onedot}
\def\tb{Table}\def\tbs{Tables}
\def\model{SpliceMix}
\def\modelExt{SpliceMix-CL}

\makeatletter
\newcommand{\removelatexerror}{\let\@latex@error\@gobble}
\newcommand\upA[1]{(\textcolor{purple}{#1$\uparrow$})}
\newcommand\dnA[1]{(\textcolor{teal}{#1$\downarrow$})}
\makeatother

\begin{document}

\title{\model: A Cross-scale and Semantic Blending Augmentation Strategy for Multi-label Image Classification}

\author{Lei Wang,
      Yibing Zhan,
      Leilei Ma,
      Dapeng Tao,
      Liang Ding,
      and Chen Gong,~\IEEEmembership{Member,~IEEE}
      \thanks{L. Wang and C. Gong are with the Key Laboratory of Intelligent Perception and Systems for High-Dimensional Information of Ministry of Education, Jiangsu Key Laboratory of Image and Video Understanding for Social Security, School of Computer Science and Engineering, Nanjing University of Science and Technology, Nanjing 210094, Jiangsu, China (e-mail: \{lei\_wang, chen.gong\}@njust.edu.cn).}
      \thanks{Y. Zhan and L. Ding are with the JD Explore Academy, Beijing 100000, China (e-mail: zhanyibing@jd.com; liangding.liam@gmail.com).}
      \thanks{L. Ma is with the School of Computer Science and Technology, Anhui University, Heifei 230601, Anhui, China (e-mail: xiaoleilei1990@gmail.com).}
      \thanks{D. Tao is with the FIST LAB, School of Information Science and Engineering, Yunnan University, Kunming 650091, Yunnan, China (e-mail: dapeng.tao@gmail.com).}
      %%%% \thanks{L. Ding is with the Shanghai Institute for Advanced Study, Zhejiang University, Shanghai 200000, China. (e-mail: liangding.liam@gmail.com)}
      \thanks{(Corresponding author: Chen Gong.)}
      }
% \author{IEEE Publication Technology,~\IEEEmembership{Staff,~IEEE,}
        % <-this % stops a space
% \thanks{This paper was produced by the IEEE Publication Technology Group. They are in Piscataway, NJ.}% <-this % stops a space
% \thanks{Manuscript received April 19, 2021; revised August 16, 2021.}}

% The paper headers
\markboth{Journal of \LaTeX\ Class Files,~Vol.~x, No.~x, November 2023}%
{Shell \MakeLowercase{\textit{et al.}}: A Sample Article Using IEEEtran.cls for IEEE Journals}

% \IEEEpubid{0000--0000/00\$00.00~\copyright~2023 IEEE}
% Remember, if you use this you must call \IEEEpubidadjcol in the second
% column for its text to clear the IEEEpubid mark.
\maketitle

\begin{abstract}
Recently, Mix-style data augmentation methods (\eg, Mixup and CutMix) have shown promising performance in various visual tasks. However, these methods are primarily designed for single-label images, ignoring the considerable discrepancies between single- and multi-label images, \ie, a multi-label image involves multiple co-occurred categories and fickle object scales.
   % co-occurrence relationships of different categories, fickle object scales.
   % In practice, multi-label images contain more complicated semantic information with multiple co-occurred objects. Existing Mix-style methods may be less beneficial for multi-label image classification (MLIC). 
   On the other hand, previous multi-label image classification~(MLIC) methods tend to design elaborate models, bringing expensive computation. 
   In this paper, we introduce a simple but effective augmentation strategy for multi-label image classification, namely \model. The ``splice'' in our method is two-fold: \emph{1)} Each mixed image is a splice of several downsampled images in the form of a grid, where the semantics of images attending to mixing are blended without object deficiencies for alleviating co-occurred bias; \emph{2)} We splice mixed images and the original mini-batch to form a new \model ed mini-batch, 
   % instead of dropping the latter, 
   which allows an image with different scales to contribute to training together. Furthermore, such splice in our \model ed mini-batch enables interactions between mixed images and original regular images. We also offer a simple and non-parametric extension based on consistency learning (\modelExt) to show the flexible extensibility of our \model. Extensive experiments on various tasks demonstrate that only using \model\ with a baseline model~(\eg, ResNet) achieves better performance than state-of-the-art methods.
   % verify the effectiveness of our proposed methods compared to state-of-the-art methods. 
   Moreover, the generalizability of our \model\ is further validated by the improvements in current MLIC methods when married with our SpliceMix. The code is available at {\textit{https://github.com/zuiran/SpliceMix}}.
\end{abstract}


\section{Introduction}
\label{sec1}
\IEEEPARstart{A}{s} a fundamental task of computer vision, image classification (especially with single-label) has been well-studied from various aspects, such as network architectures~\cite{he2016deep,mellor2021neural}, data augmentation strategies~\cite{zhang2018mixup,10288408}, and pre-trained models~\cite{he2022masked,yang2023effective}. However, some methods designed for single-label images are not always adaptive to the multi-label image classification~(MLIC) task since a multi-label image usually contains multiple categories, which means more complicated scene and brings new challenges~\cite{zhu2021residual,liu2022contextual,you2020cross}, \ie, label dependency learning and small object recognition. 
% fickle object scales, label dependency, \etc. 
In view of the fact that almost no data augmentation methods are especially designed for multi-label images according to their characters, it drives us to seek a simple but effective MLIC-adaptive augmentation strategy.

\begin{figure}[t]
   \centering  
   \subfigure[Regular Mini-batch]{
      \includegraphics[width=0.48\linewidth]{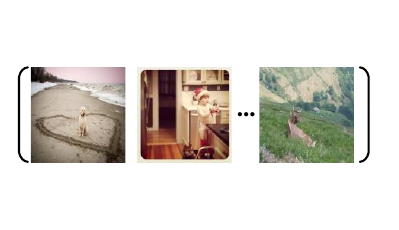}}
   \hfill %\bigskip
   \subfigure[Mix-style Mini-batch~\cite{zhang2018mixup,yun2019cutmix}]{
      \includegraphics[width=0.48\linewidth]{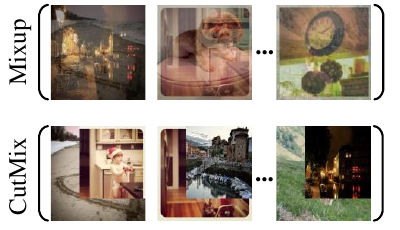}}

   % \vspace{-1.5\baselineskip}
   \subfigure[BA Mini-batch~\cite{hoffer2020augment}]{
      \includegraphics[width=0.48\linewidth]{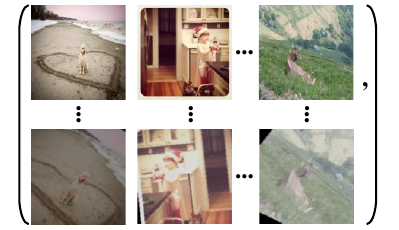}}
   \hfill %\bigskip
   \subfigure[\model ed Mini-batch]{
      \includegraphics[width=0.48\linewidth]{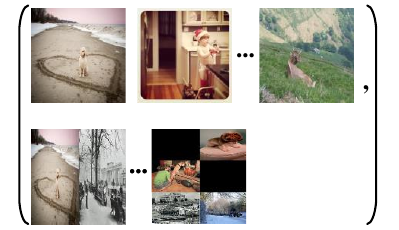}}

   % \vspace{-1.1\baselineskip}
   \caption{Mini-batch comparisons between our \model\ and other methods. Our \model\ augments the image and label space similar to (b), and batch scale similar to (c), leading to good generalizability and fast convergence.}      
   \label{fig1.1}
\end{figure}

\par
% Deep learning-based MLIC methods typically transform multi-label classification problem to per-class binary classification problem in a fashion of problem transformation~\cite{zhang2013review}.
% \sout{, like Binary Relevance~\cite{boutell2004learning}} and utilize binary cross entropy (BCE) rather than categorical cross entropy (CCE or CE for simplicity) as the objective loss. 
% Deep learning-based MLIC methods typically transform multi-label classification problem to per-class binary classification problem in a fashion of problem transformation~\cite{zhang2013review}, like Binary Relevance~\cite{boutell2004learning}, and then utilize binary cross entropy (BCE) rather than categorical cross entropy (CCE or CE for simplicity) as the objective loss. 
In MLIC, label dependency has been widely studied in \cite{ye2020attention,chen2019multi,lanchantin2021general,deng2022beyond,zhou2023mining}, which builds the co-occurred relationship of different categories for boosting recognition performance.
 % of co-occurred categories. 
 Nonetheless, as claimed in \cite{singh2020don,liu2022contextual}, the label co-occurrence pattern learned from the training set could be contextually biased to identify a category when its typical context is absent. On the one hand, the semantic context~(\ie, co-occurred relationship of categories) is crucial for improving MLIC. On the other hand, it may mislead the model in the scene of a category occurring in its unseen context. It is tricky to make a trade-off between learning label dependency and reducing co-occurred bias.
% Apart from the exclusive issue of contextual bias~\cite{singh2020don} in MLIC, background bias~\cite{xiao2020noise} also hinders a model's generalizability no matter in single- or multi-label image classification. We follow the terminology of scene bias introduced in~\cite{mo2021object} to represent both contextual bias (\ie, co-occurred bias of different categories) and background bias (\ie, co-occurred bias of background and categories). Although some methods~\cite{singh2020don,xiao2020noise,mo2021object} are proposed for dealing with these biases, some of them focus on single-label images and some of them focus on the situation of extreme contextual bias that loses the practicability for the common MLIC task. Also, they are all dependent on well-designed models that are lack of efficiency and flexibility.
% Another plain is 
\par
To some extent, Mix-style data augmentation methods naturally reduce the co-occurred bias. They usually generate a new image via mixing two images and their labels, which randomly produces a context-agnostic scene while losing the co-occurred relationship of categories. Two representative Mix-style methods, Mixup~\cite{zhang2018mixup} and CutMix~\cite{yun2019cutmix}, have drawn lots of attention and their variants~\cite{kim2020puzzle,liu2022tokenmix,olsson2021classmix} have shown huge potential for many visual tasks recently. However, these methods are mainly designed for single-label images, neglecting the considerable gap between single- and multi-label images.
% , \eg, fickle object scales presented in the same image and more complicated image semantics with multiple co-occurred objects, are neglected. 
As aforementioned, they fail to capture label dependency. Moreover, the small objects are easy to be messed in Mixup or lost in CutMix.
% The small objects are easy to be messed in Mixup or lost in CutMix. Moreover, a linearly weighted combination of two multi-label images in Mixup may be too complex for a model to learn valuable representation \rv{from} a multi-label image. 
Although some variants extract salient object regions of an image to paste to another image that may not suffer from the above pains, we want to claim that these extracted regions could be unreliable and they require additional knowledge~\cite{su2021context} or complicated design~\cite{kim2020puzzle,uddin2020saliencymix}.

\par
In this paper, we introduce a simple but effective, MLIC-adaptive Mix-style augmentation strategy, namely \model. The proposed method is not only targeted on MLIC, but also tackles another well-known issue of Mix-style methods that requires longer training epochs~\cite{zhang2018mixup,yun2019cutmix}. 
% where they usually need longer training time for reaching satisfying performance~\cite{zhang2018mixup,yun2019cutmix}. 
Inspired by the fast convergence of Batch Augmentation~(BA)~\cite{hoffer2020augment},
% that augments the mini-batch via replicating samples with different transformations repetitively, 
we keep both original regular mini-batch and generated samples to form a new \model ed mini-batch. \fig~\ref{fig1.1} compares the mini-batch used in regular training, Mix-style methods, BA, and \model . 
% The comparison of a mini-batch used in regular training, Mix-style methods, BA and \model\ is illustrated in \fig~\ref{fig1.1}. 
Comparing \figs~\ref{fig1.1}~(c) and (d), we show that \model\ increases the mini-batch size with only several mixed images, which is far less than batch scale of BA. Because each generated image is a splice of multiple images, a little increase of batch size is enough to guarantee fast convergence. Comparing mixed samples in \figs~\ref{fig1.1}~(b) and (d), we show that unlike existing Mix-style methods, our \model\ blends image semantics while preserving the entire objects, which produces an unusual scene for potential co-occurrences of both different objects and backgrounds-objects.
In brief, our \model\ augments the sample space and batch scale simultaneously, where the former is to blend image semantics for alleviating co-occurred bias, and the latter allows us to learn label dependency from regular images and leads to fast convergence. A balance between learning label dependency and reducing co-occurred bias can be obtained in our \model. 
% Comparing \figs~\ref{fig1.1}~(c) and (d), we show that \model\ increases the mini-batch size with only several mixed images, which is far less than batch scale of BA. Because each generated image is a splice of multiple images, a little increase of batch size is enough to guarantee fast convergence of testing error. Comparing mixed samples in \figs~\ref{fig1.1}~(b) and (d), we show that unlike existing Mix-style methods, our \model\ blends image semantics while preserving the entire objects, which produces an unusual scene for potential co-occurrences of both different objects and background-objects. 
As shown in \fig~\ref{fig1.1}~(d) and \fig~\ref{fig3.1}, the mixed image in the form of a grid of multiple downsampled images contributes to cross-scale training with regular images, beneficial to small object recognition.
% our \model\ generates a new image in the form of a grid of multiple downsampled images. An image can have multiple scales in a \model ed batch, which results in cross-scale training and contributes to small object recognition. 
Furthermore, the splice of regular images and mixed images enables the interactions between the two, which permits flexible extensions of \model. Hence, we offer a non-parametric design based on consistency learning, \ie, \modelExt, to show the extensibility of our \model. 

\par
Our contributions can be summarized as:
\begin{itemize}
    \item To our best knowledge, our \model\ is the first data augmentation strategy designed for MLIC. The proposed \model\ is not only simple yet effective, but also orthogonal to existing MLIC methods, which can boost these methods remarkably. 
    % almost no data augmentation methods designed for MLIC, we introduce a simple but effective augmentation strategy, namely \model, for the MLIC task. 
    % The proposed \model\ is orthogonal to existing MLIC methods, which can boost these methods remarkably.
    \item We offer a non-parametric, consistency learning-based extension to show the flexible extensibility of \model, where the fine knowledge of regular images is leveraged to learn better representation for mixed images.
    \item The two proposed methods achieve superior performance on several popular MLIC tasks and data sets. We also conduct on comprehensive analysis experiment to clarify the effectiveness of the proposed \model.
\end{itemize}

\section{Related Works}
\label{sec2}
\subsection{Multi-label Image Classification.}
Multi-label image classification (MLIC) is a challenging computer vision task and has attracted lots of attention. Existing MLIC methods prefer designing elaborate models to capture attention regions or label dependencies.

\par
Discovering attention regions in a multi-label image is helpful to predict present categories or generate proposal candidates. Spatial Regularization Network (SRN)~\cite{zhu2017learning} learns weighted spatial attention maps with only image-level supervision and used multi-layer convolution to estimate class confidences from such attention maps. 
% \cite{huynh2020shared} proposes class-agnostic shared attention for zero-shot learning. 
Different from SRN, \cite{zhu2021residual} reuses classifier weights to obtain class-specific attention for discovering spatial discriminative regions. To generate informative proposal candidates, \cite{chen2018recurrent,gao2021learning,zhan2022global} utilize spatial attention to locate object regions. The difference among them is that \cite{chen2018recurrent,zhan2022global} extract local region features from feature maps directly instead of feeding local image regions into CNN again used in~\cite{gao2021learning}.

\par
For label dependency-based methods, some techniques, \eg, Recurrent Neural Network (RNN)~\cite{zaremba2014recurrent}, Graph Neural Network (GNN)~\cite{kipf2016semi} and Transformer~\cite{vaswani2017attention}, usually are utilized to model label co-occurred correlation. CNN-RNN~\cite{wang2016cnn} learns joint image-label embedding and builds high-order label dependency via long short term memory recurrent neurons~\cite{narayan2021discriminative}. 
% While, the label order is required to predefined in CNN-RNN for both training and testing. \cite{chen2018order,yazici2020orderless} propose order-free methods to tackle this problem. 
Beneficial from the excellent ability of modeling correlation GNN possesses, a series of methods~\cite{chen2019multi,zhou2023mining,deng2022beyond} are proposed. \cite{chen2019multi,chen2019learning} consider label co-occurrence statistics as the adjacency matrix and then exploit GNN to map the label graph to class-dependent classifier or label embeddings. In view of poor model generalizability of label statistics-based graph, \cite{ye2020attention,zhao2021transformer} introduce dynamic graph to eliminate co-occurrence bias from the training set. In \cite{lanchantin2021general}, Transformer Decoder is utilized to learn complex dependencies between feature maps and label embeddings. \cite{liu2021query2label,zhu2022two} leverages both Encoder and Decoder in Transformer to learn intra-feature and feature-label correlations.

\par
Two recent works~\cite{liu2022contextual,liu2023causal} indicate that exploiting label dependency may cause contextually biased recognition. Coincidentally, some previous MLIC researches~\cite{zhu2021residual,ye2020attention} suggest that the label co-occurrence information from limited training data is not enough for MLIC, which could lead to over-fitting. Therefore, we propose the \model\ to mitigate co-occurred bias via generating unusual scenes where the semantics of several images are blended and the learned label co-occurrence pattern is ameliorated.

\begin{figure*}[!h]
    \centering
    \includegraphics[width=1\linewidth]{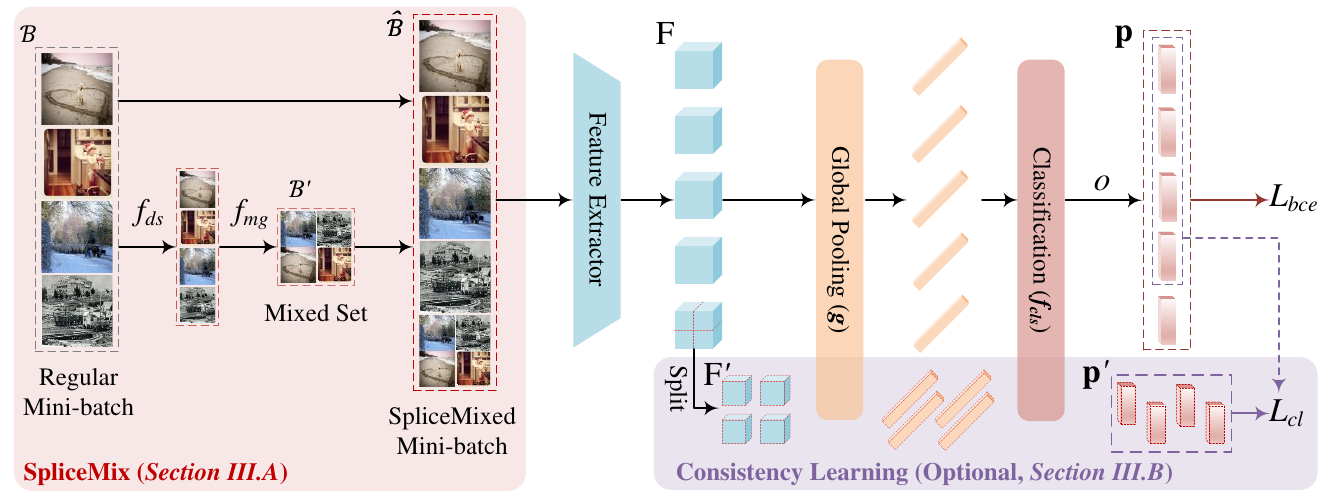}
    \caption{Overview of \model\ and \modelExt\ with a $2\times 2$ grid strategy. ``$\stackrel{\bullet}{\longrightarrow}$'' denotes a function operation and ``\textcolor{violet}{$\dashrightarrow$ }'' denotes calculation without backpropagation. We simply introduce \model\ with the mini-batch samples whose size is 4.  For generating unusual scene and preserving intact objects, \model\ makes a grid of downsampled images randomly sampled from the regular mini-batch to form the mixed set. The final \model ed mini-batch is obtained by combining the regular mini-batch and mixed set. Furthermore, we offer a consistency learning-based design to show the flexible expansibility of our \model.}      
    \label{fig3.1}
\end{figure*}

\subsection{Mix-style Augmentation.}
Recently, Mix-style augmentation methods have shown promising performance in various single-label visual tasks, such as image classification~\cite{yun2019cutmix,liu2022tokenmix}, super-resolution~\cite{yoo2020rethinking} and semantic segmentation~\cite{olsson2021classmix}.
% Mix-style methods combine multiple data to generate new images, augmenting the space of both images and labels. 
As a founding one, Mixup~\cite{zhang2018mixup} generates a new image by linearly combining two images and their labels. Since then, a bunch of Mix-style methods have been proposed, where CutMix~\cite{yun2019cutmix} is a representative work that joints Mixup and Cutout~\cite{devries2017improved} via replacing local regions with a patch from another image.  Similar to the way of cut-and-paste in CutMix, some methods insert new patches to target images via leveraging smooth transition\cite{lee2020smoothmix}, attention maps~\cite{walawalkar2020attentive}, saliency information~\cite{uddin2020saliencymix,kim2020puzzle}. Considering limited benefits of CutMix for visual Transformers, \cite{liu2022tokenmix} mixes two images at token-level and generates the mixed label from content-based activations of a teacher network. \cite{yoo2020rethinking} extends CutMix to the image super-resolution task where images with high resolution and low resolution consist of sample-pairs. \cite{olsson2021classmix} exploits predicted object boundaries to mix two unlabeled images for semi-supervised semantic segmentation. 

Although Mix-style methods behave well in diverse single-label visual tasks, they are not suitable to MLIC because of more complicated image semantic and more categories in an multi-label image. For instance, cut-and-paste methods fail to assign the weight of labels from two images since a local patch cannot represent a multi-label images. In view of this, we aim to design a MLIC-adaptive augmentation strategy and exploit entire images to attend to mixing, which preserves all objects and results in reliable mixed labels.

% Although these Mix-style methods have made a great progress, they usually require longer training epochs to ensure convergence.

% Similar to the way of cut-and-paste in CutMix, some methods insert new patches to target images via leveraging smooth transition\cite{lee2020smoothmix}, attention maps~\cite{walawalkar2020attentive}, saliency information~\cite{uddin2020saliencymix,kim2020puzzle}. Considering limited benefits of CutMix for visual Transformers, \cite{liu2022tokenmix} mixes two images at token-level and generates the mixed label from content-based activations of a teacher network. \cite{yoo2020rethinking} extends CutMix to the image super-resolution task where images with high resolution and low resolution consist of sample-pairs. \cite{olsson2021classmix} exploits predicted object boundaries to mix two unlabeled images for semi-supervised semantic segmentation. Although Mix-style methods behave well in diverse single-label visual tasks, they are not suitable to MLIC because of more complicated image semantic and more categories in an multi-label image. For instance, cut-and-paste methods fail to assign the weight of labels from two images since a local patch cannot represent a multi-label images. In view of this, we aim to design a MLIC-adaptive augmentation strategy and exploit entire images to attend to mixing, which preserves all objects and results in reliable mixed labels. \textcolor{teal}{\large{\textbf{[Mosaic]} to do?}}

% \par
\subsection{Batch Augmentation.}
Different from previous augmentation methods, Batch Augmentation~(BA)~\cite{hoffer2020augment} increases the mini-batch size by replicating samples within the same mini-batch with different data augmentations, which accelerates model convergence and shows good generalization. The original BA is proposed to improve performance of training with large mini-batch size. \cite{fort2021drawing} empirically shows that BA also works well for small mini-batch size. Whereas, BA still requires augmenting each image repeatedly for building a mini-batch, which is demanding. In our \model, we slightly increase the mini-batch size with mixed samples, each of which consists of multiple downsampled regular images. Even with only a few mixed samples, \model\ can significantly boost performance. Same to \cite{fort2021drawing}, we also allow fixed mini-batch size, \ie, keeping the same size to regular mini-batch via reducing regular samples to save space for mixed images. 

% Strictly speaking, Batch Augmentation~(BA)~\cite{hoffer2020augment} is not a specific data augmentation method, because it dose not enlarge sample scale. BA increases the batch size by replicating samples within the same batch with different data augmentations, which accelerates model convergence and shows good generalization. Hence, we consider BA as a special type of data augmentation that enlarges sample tuples each of which contains the same image with different transformations in the view of a mini-batch, compared with only one image per sample tuple in regular training. The original BA is proposed to improve performance of training with large batch size. \cite{fort2021drawing} empirically shows that BA also works well for small batch size. Whereas, BA still requires to augment each image repeatedly for building a mini-batch, which is demanding. In our \model, we slightly increase the mini-batch size with mixed samples that consists of multiple downsampled regular images. The proposed method even requires merely a mixed sample whiles boost performance obviously. Same to \cite{fort2021drawing}, we also allow fixed mini-batch size, \ie, keeping the same size to regular mini-batch via reducing regular samples to save space for mixed images.

\section{Methodology}
\label{sec3}
This section will detail the proposed \model. Benefiting from our generated \model ed mini-batch that enables interactions between original samples and their mixed samples, we also give a non-parametric extension based on consistency learning, namely \modelExt, for further improvement.

\subsection{\model}
\label{sec3.1}
Given a mini-batch $\mathcal{B}=\{(\textbf{x}_i, \textbf{y}_i)|i\in[B]\}$, randomly sampling $B$ points from the training set, where $\textbf{x}$ and $\textbf{y}$ are an image and its label, respectively. As illustrated in \fig~\ref{fig3.1}, \model\ will generate a new mixed sample set $\mathcal{B}'$ from $\mathcal{B}$, and then concatenate the original mini-batch and generated set to form the final mini-batch $\hat{\mathcal{B}}$ for training, \ie, $\hat{\mathcal{B}} = \mathcal{B} \cup \mathcal{B}'$. 
% Mathematically, that is $\hat{\mathcal{B}} = \mathcal{B} \cup \mathcal{B}'$. 
Before turning sights to $\mathcal{B}'$, we would like to clarify some terms used in the rest to avoid confusion. We call the original mini-batch $\mathcal{B}$ as the regular batch whose samples $(\textbf{x}$, $\textbf{y})$s are regular ones, the new mixed sample set $\mathcal{B}'$ as the mixed set whose samples $(\textbf{x}'$, $\textbf{y}')$s are mixed ones and the generated final mini-batch $\hat{\mathcal{B}}$ as the \model ed batch.

\textbf{Mixture of images:}
% For a multi-label image, multiple objects from various categories may attend concurrently. Previous CutMix-like methods~\cite{liu2022tokenmix,chen2022transmix} usually assume that local regions can be in charge of the whole image whose label contributes to mixing directly. However, we claim that such way is not adaptive to the MLIC task since a random region in a multi-label image fails to represent the whole image, which is more impossibly in charge of one concrete category. 
The proposed \model\ aims to generate new images, preserving complete objects and producing an unusual scene for present categories. To achieve this, we make a grid of chosen regular images from $\mathcal{B}$ to form the mixed image. Keeping the resolution consistent with regular images, the chosen images are downsampled before griding, which brings a benefit of enabling the model to learn multi-scale information from regular images and their mixed ones. Let $f_{mg}$ denote the operation of making a grid and $f_{ds}$ denote the operation of downsampling. We can express the mixed image as 
% A mixed image can be obtained by
\begin{equation}\label{eq3.1}
    \textbf{x}' = f_{mg}(\{f_{ds}(\textbf{x}_i)|i\in\Omega\})~,
\end{equation}
where $\Omega$ is an index set of the regular batch that accounts for randomly selecting regular images to form the mixed image, \eg, the cardinality of $\Omega$ will be 4 if we want to obtain a mixed image with a $2\times 2$ grid of regular images. 
% For as possibly and efficiently as utilizing samples in the entire regular batch, the chosen images are unique for each mixed image, \ie, for the mixed set $\mathcal{B}'=\{\textbf{x}'_j|j\in[M]\}$, $\cap_{j\in[M]}\Omega_j=0$, where $M=|\mathcal{B}'|$ is a hyper-parameter to adjust the magnitude of $\mathcal{B}'$. \textcolor{teal}{Note that there might be no performance discrepancy if we generate the mixed set from partially repeated regular images with larger $M$, which we will discuss in \color{green}{Experiments}.} Since we want to offer a simple while both effective and efficient strategy, we suggest to fix $M$ according to the grid number and regular batch size.

\par
In terms of the mixed images, they are a little similar to Mosaic used since YOLOv3~\cite{redmon2018yolov3}. Here, we claim that the two methods are different in three points: \emph{1)} Mosaic needs bounding boxes of objects for annotation that cannot be applied to MLIC; \emph{2)} Objects in vanilla Mosaic are in original scale, not involved to cross-scale learning; \emph{3)} The mixed set is just a part of our \model\ where the regular images are kept, and both usual scenes (from regular images) and unusual scenes (from mixed images) can be learned.

\textbf{Mixture of labels:}
Different from existing Mix-style methods that mix labels of two images via linear combination, the proposed \model\ sets the class label to be true if this class is present in the mixed image, formulated by 
\begin{equation}\label{eq3.2}
    \textbf{y}' = \cup_{i\in\Omega}\textbf{y}_i~.
\end{equation}
The mixed label keeps multi-hot same to regular samples while containing more categories, which enriches the label diversity and is more suitable for MLIC training than the soft label used in existing Mix-style methods. 
Our \model\ method can be implemented readily in several lines of code. A PyTorch~\cite{paszke2019pytorch} implementation of $2\times 2$ mixed strategy is presented in \fig~\ref{fig3.2}.
% , \textcolor{teal}{that is our main solution adopted for experiments due to a good trade-off it possesses between classification performance and computation budget, although there exist better choices for higher recognition with a little more computation.}

\par
Since we use the global images rather than local regions for mixing, there is almost no information depletion of present objects (except their resolutions are reduced) that means the label of each mixed image is reliable. More importantly, the proposed \model\ puts objects into a rare or complex scene, \ie, mixed scenes in our generated image. The model will build correlations among objects who may be co-occurred unusually and between objects and backgrounds where objects may be seldomly present. The mixed samples blend the image semantics and give the model more chances to see these potential scenes, alleviating the semantic and contextual bias and enhancing the model's generalizability. On the other hand, we agree with the importance of inherent correlations from the training set and preserve the regular batch in our \model ed batch in a BA-like way. The correlations on highly co-occurred objects still can be learned. In brief, our \model ed batch consists of the regular batch and mixed set for the training time, from the former of which common label correlations are built and from the latter uncommon label correlations are reinforced. Both of them boost the final recognition and generalization performance.

\begin{figure}[t]
    \centering
    % \begin{minipage}{.55\linewidth}
        \RestyleAlgo{boxed}
        \SetAlFnt{\footnotesize\ttfamily} % ttfamily rmfamily
        % \SetAlgoNlRelativeSize{-.5}
        % \SetNlSkip{.5em}
        % \SetNlSty{}{}{}
        \removelatexerror
        \begin{algorithm}[H]\label{alg1}
            \textrm{import random, torch}  \\
            \textrm{from torch.nn.functional import \textcolor{teal}{interpolate} as \textcolor{teal}{f\_ds}}  \\
            \textrm{from torchvision.utils import \textcolor{teal}{make\_grid} as \textcolor{teal}{f\_mg}}  \\
            \ \\
            \textcolor{cyan}{def} \textcolor{Melon}{\model}(X, Y):  \\
            \textcolor{CadetBlue}
            {\#~X:~(\textbf{B}atch size, \textbf{C}hannels, \textbf{H}eight, \textbf{W}idth)  \\
             \#~Y:~(Batch size, classes)}\\
            \quad\quad B, C, H, W \textcolor{purple}{=} X.\textcolor{teal}{shape}  \\
            \quad\quad Omega \textcolor{purple}{=} random.\textcolor{teal}{sample}(\textcolor{teal}{range}(B), B\textcolor{purple}{//}4\textcolor{purple}{*}4)  \\

            \quad\quad X\_ds \textcolor{purple}{=} \textcolor{teal}{f\_ds}(X[Omega], \textcolor{olive}{size}\textcolor{purple}{=}(H\textcolor{purple}{//}2, W\textcolor{purple}{//}2),  \\
            \quad\quad\quad\quad\textcolor{olive}{mode}\textcolor{purple}{=}'bilinear', \textcolor{olive}{align\_corners}\textcolor{purple}{=}True) \\

             \quad\quad X\_ \textcolor{purple}{=} \textcolor{teal}{f\_mg}(X\_ds, \textcolor{olive}{nrow}\textcolor{purple}{=}2, \textcolor{olive}{padding}\textcolor{purple}{=}0)  \\

            \quad\quad X\_ \textcolor{purple}{=} X\_.\textcolor{teal}{split}(H, \textcolor{olive}{dim}\textcolor{purple}{=}1)  \\
            \quad\quad X\_ \textcolor{purple}{=} torch.\textcolor{teal}{stack}(X\_, \textcolor{olive}{dim}\textcolor{purple}{=}0)  \\
            \quad\quad Y\_ \textcolor{purple}{=} Y[Omega].\textcolor{teal}{view}(B\textcolor{purple}{//}4, 4, -1).\textcolor{teal}{sum}(1)  \\ % 
            \quad\quad Y\_[Y\_\textcolor{purple}{\textgreater}0] \textcolor{purple}{=} 1  \\

            \ \\
            \quad\quad X\_hat \textcolor{purple}{=} torch.\textcolor{teal}{cat}((X, X\_), \textcolor{olive}{dim}\textcolor{purple}{=}0)  \\
            \quad\quad Y\_hat \textcolor{purple}{=} torch.\textcolor{teal}{cat}((Y, Y\_), \textcolor{olive}{dim}\textcolor{purple}{=}0)  \\
            \quad\quad \textcolor{cyan}{return} X\_hat, Y\_hat
        \end{algorithm}
    % \end{minipage}
    \caption{PyTorch implementation of \model\ with a $2\times 2$ grid strategy.}
    \label{fig3.2}
\end{figure}

% \par  It is a pity to omit.
% Since we use the global images rather than local regions for mixing, there is almost no information depletion of present objects (except their resolutions are reduced) that means the label of each mixed image is reliable. More importantly, the proposed \model\ put objects into a rare or complex scene, \ie, mixed scenes in our generated image. The model will build correlations among objects who may be co-occurred unusually and between objects and backgrounds where objects may be seldomly present. The mixed samples blend the image semantics and give the model more chances to see these potential scenes, alleviating the semantic and contextual bias and enhancing the model's generalizability. On the other hand, we agree with the importance of inherent correlations from the training set and preserve the regular batch in our \model ed batch in a BA-like way. The correlations on highly co-occurred objects still can be learned. In brief, our \model ed batch consists of the regular batch and mixed set for the training time, from the former of which common label correlations are built and from the latter uncommon label correlations are reinforced. Both of them boost the final recognition and generalization performance.

\textbf{Setting of grids:}
As stated in \eq.~\eqref{eq3.1}, a mixed image is made from chosen regular images via two operations, \ie, $f_{mg}$ and $f_{ds}$, where the downsampling operation $f_{ds}$ is to resize the chosen regular images that lets the resolution of a mixed image match with that of regular images. For instance, given 4 regulars image whose resolutions are $448\times 448$ for generating a mixed image with a $2\times 2$ grid, we firstly resize them to $224\times 224$ and then make a $2\times 2$ grid of them to form the mixed image. The reduced scale $f_{ds}$ depends on which kind of the grid used for generating a mixed image. 
The grid number of different mixed images can be various. In other words, our \model\ allows to mix images with different combinations of grids to make up a mixed set, such as mixed images with grids of $1\times 2$, $2\times 2$, $2\times 3$, \etc. 

\par
\begin{figure}[t]
   \centering  
   \subfigure[Sym. grids]{
      \begin{minipage}{0.2\linewidth}
         \centering
         \includegraphics[width=1\textwidth]{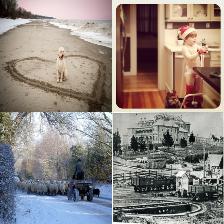}
         \par
         {\fontsize{8}{10}\selectfont $2\times 2$}
         \vspace{+.3\baselineskip}
      \end{minipage}
      % \hfill
      % \begin{minipage}{0.15\linewidth}
      %    \centering
      %    \includegraphics[width=1\textwidth]{fig3_3/grid_(3, 3).jpg}
      %    \par
      %    {\fontsize{8}{10}\selectfont $3\times 3$}
      % \end{minipage}
   }
   \hfill
   \subfigure[Asym. grids]{
      \begin{minipage}{0.2\linewidth}
         \centering
         \includegraphics[width=1\textwidth]{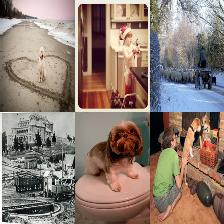}
         \par
         {\fontsize{8}{10}\selectfont $2\times 3$}
         \vspace{+.3\baselineskip}
      \end{minipage}
      \hfill
      \begin{minipage}{0.2\linewidth}
         \centering
         \includegraphics[width=1\textwidth]{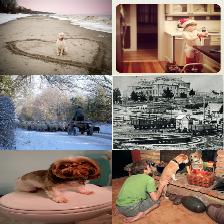}
         \par
         {\fontsize{8}{10}\selectfont $3\times 2$}
         \vspace{+.3\baselineskip}
      \end{minipage}
   }
   \hfill
   \subfigure[Grids~w/~dp]{ % 
      % \begin{minipage}{0.15\linewidth}
      %    \centering
      %    \includegraphics[width=1\textwidth]{fig3_3/grid_(2, 2)_bk_1.jpg}
      %    \par
      %    {\fontsize{8}{10}\selectfont $2\times 3$}
      % \end{minipage}
      % \hfill
      \begin{minipage}{0.2\linewidth}
         \centering
         \includegraphics[width=1\textwidth]{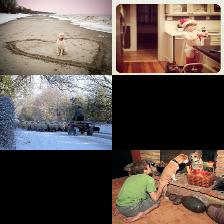}
         \par
         {\fontsize{8}{10}\selectfont $3\times 2-2$}
         \vspace{+.3\baselineskip}
      \end{minipage}
   }
   % \subfigure[$2\times 2$]{
   %    \begin{minipage}{0.48\linewidth}
   %       \centering
   %       \includegraphics[width=1\textwidth]{fig3_3/grid_(2, 2).jpg}
   %       % \caption{$2\times 2$}
   %    \end{minipage}
   %    \hfill
   %    \begin{minipage}{0.48\linewidth}
   %       \centering
   %       \includegraphics[width=1\textwidth]{fig3_3/grid_(3, 3).jpg}
   %       % \caption{$2\times 2$}
   %    \end{minipage}
   % }

   % \vspace{-2\baselineskip}
   \caption{Illustration of several feasible grid strategies for \model. Firstly, we assume that the regular images are square. Then, we can summary these grids as three types: (a) Symmetric grids with the same rows and columns; (b) Asymmetric grids with distinct rows and columns; (c) Grids with dropout where some sub-images are discarded randomly.}  
    % where (a) partly lists grids with the same columns and rows, (b) partly lists grids with different columns and rows, and (c) presents mixed images whose sub-images are dropped randomly and replaced with black, \eg, the left of (c) drops a sub-image from the $2\times 2$ mixed image.    
   \label{fig3.3}
\end{figure}

Here, we focus on the grid strategy introduced for \model. Several feasible grids are illustrated in \fig.~\ref{fig3.3}. Accordingly, we can define a function family $\mathcal{F}_{mg}$ for sampling $f_{mg}$, that is
\begin{equation}\label{eq3.3}
    \mathcal{F}_{mg}=\{f^{r\times c}|r\in[G_r],c\in[G_c]\}~,
\end{equation}
where we rewrite $f_{mg}$ to $f^{r\times c}$, an operation of making a $r\times c$ grid, and $G_r$, $G_c$ denote the maximal rows and columns, respectively. Next, three types of the grid shown in \fig.~\ref{fig3.3} are discussed:
\begin{enumerate}[label=\textbf{\arabic*)}]
    \item \textbf{Symmetric grids:} Sub-images in a symmetric grid are proportionally scaled from their regular images. Due to the same columns and rows in a symmetric grid, $f^{r\times c}$ is equal to $f^{c\times r}$. The number of grid dominates the complexity of the generated image. A more complicated mixed image needs more regular images whose resolution will be downsampled much more.
    \item \textbf{Asymmetric grids:} The shape of sub-images in an asymmetric grid is squashed compared with their regular ones, 
    which can be viewed as an additional transform besides rescaling. For a $f^{r\times c}$, if we keep the columns $c$ larger than the rows $r$ (\eg, the left figure in \fig~\ref{fig3.3}~(b)), the model may learn to recognize stretched objects well while losing its generalizability to flatten objects. To avoid this, we throw a coin with a flipped probability of 0.5 to decide which of $\{f^{r\times c},~f^{c\times r}\}$ will be adopted under this setting.
    % To avoid learning biases from distorted objects, we throw a coin with a flipped probability of 0.5 to decide which of $\{f^{r\times c},~f^{c\times r}\}$ will be adopted.
    \item \textbf{Grids with dropout:} 
    % Considering that the mixed images may be over-complicated due to a large grid number, we exploit a simple dropout mechanism to randomly mask a part of sub-images and their corresponding labels in a mixed image.
    The goal of mixed images in the form of grids is to blend the image semantics and generate a new, unusual scene. Whereas, the generated scene may be over-complicated due to a large grid number, \ie too many regular images attend to a mixed image. Therefore, we exploit a simple dropout mechanism to randomly mask a part of sub-images and corresponding labels in a mixed sample. After this, the complexity of a mixed image can be reduced, meanwhile the diversity of training samples is further enhanced.  
\end{enumerate}

% In practice, the large discrepancy between the resolutions of training and testing time harms the recognition performance~\cite{touvron2019fixing}. In \model, the resolution discrepancy exists in both training time and training-testing time. In training time, a mixed image consists of multiple downsampled sub-images whose area ratio to the regular images is at most 0.5 when $f^{1\times 2}$ or $f^{2\times 1}$ is adopted. The learning capacity of the model may be influenced negatively because of the large resolution difference between regular images and sub-images. Similar to testing time, the model learning from excessively small sub-images may be unhelpful to infer testing images whose size is greatly different from sub-images. \textcolor{teal}{Taking the above two points into account, we set both $G_c$ and $G_r$ in \eq~\eqref{eq3.3} to not larger than 3 for good learning and inference capacity. }

\subsection{\model-CL}
% Let us recall that the \model ed batch $\hat{\mathcal{B}}$ consists of a regular batch $\mathcal{B}$ and a mixed set $\mathcal{B}'$, \ie, $\hat{\mathcal{B}} = \mathcal{B} \cup \mathcal{B}'$. Mixed images in $\mathcal{B}'$ are generated in the form of a grid of multiple downsampled images from $\mathcal{B}$. The sub-images (\ie, downsampled, regular images) of a mixed image contain highly similar semantic information to their regular ones, because the only difference between them is the resolution. While, as mentioned before, the gap of polarized resolutions may harm the learning capacity. Besides this, the learned representation of sub-images with low resolution is coarse, compared with fine representation of regular images with high resolution.

The proposed \model\ allows interactions between sub-images from a mixed image and their regular versions, which means that the fine knowledge learned from regular images can be utilized to guide the model to learn better representation about coarse sub-images. In fact, there exist many techniques to help us learn consistent knowledge between regular images and sub-images, such as logits-based~\cite{zhang2018deep,zhao2022decoupled} or feature-based~\cite{chen2021cross,gou2021knowledge} knowledge distillation methods and class decoupling-based~\cite{chen2019learning,ye2020attention} or attention-based~\cite{ridnik2021asymmetric} MLIC methods, where the knowledge distillation-aware methods are used for consistency learning and the MLIC-aware methods are used for task-specific knowledge extraction. Although these techniques may show promising performance with our \model, an elaborate model design is beyond our current study and we leave it to future work. Here, we offer a simple, non-parametric idea based on consistency learning~\cite{hinton2015distilling}, namely \model-CL, to show the extensibility of \model.

\par
Assume that $\text{F}$ is the last feature map extracted from a feature extractor, \eg, CNNs~\cite{he2016deep,liu2022convnet}, 
% or visual Transformers~\cite{dosovitskiy2020image,li2022exploring}, 
$g$ is a global pooling operation, $f_{cls}$ is a linear classifier, $o$ is sigmoid activation. Then, we can obtain the label prediction $\textbf{p}$ of an input image, that is $\textbf{p}=o(f_{cls}(g(\text{F})))$. In a SpliceMixed batch, the classification loss, \ie, binary cross entropy~(BCE) loss, is
% Same to \cite{chen2019learning,zhu2021residual}, the binary cross entropy~(BCE) loss is adopted as the classification loss.
% The binary cross entropy~(BCE) loss of a SpliceMixed batch is
\begin{equation}
    \mathcal{L}_{bce} = \sum_{i}^{|\hat{\mathcal{B}}|}{-\textbf{y}_i log(\textbf{p}_i) - (1-\textbf{y}_i)log((1-\textbf{p}_i))}~,
\end{equation}
where $log$ denotes the element-wise logarithmic function. For convenience, we slightly abuse $(\textbf{x}, \textbf{y})$ to denote any sample from $\hat{\mathcal{B}}$ and keep vector form of loss, each element of which is a class-specific loss. A scalar form of loss can be obtained by summing all class-specific loss values.

For each sub-image in a mixed image, we obtain its feature map $\text{F}'$ via splitting $\text{F}$ according to the grid form of the mixed image (see \fig~\ref{fig3.1}). 
By the way, we can simply exploit $f_{mg}$ to recover $\text{F}$ from $\text{F}'$s: $\text{F}=f_{mg}(\{\text{F}'_i|i\in\Omega\})$, where the subscript $i$ indicates the $i$-th image in $\mathcal{B}$ attending to the mixed image and we reuse such subscript to indicate the feature map of a sub-image since a regular image attends to mixed images at most once. 
Consequently, the prediction $\textbf{p}'$ of a sub-image can be calculated by $o(f_{cls}(g(\text{F}')))$. For consistency learning between sub-images and their regular images, we utilize the prediction distribution \textbf{p} from a regular image to supervise the output of its corresponding sub-image, whose objective can be formulated by

\begin{equation}\label{eq3.5}
    \mathcal{L}_{cl} = \sum_{i}^{|\mathcal{B}'|}\sum_{j\in\Omega_i}{-\bar{\textbf{p}}_j log(\textbf{p}'_j) - (1-\bar{\textbf{p}}_j)log(1-\textbf{p}'_j)}~,
\end{equation}
where $\bar{\textbf{p}}$ is a copy of $\textbf{p}$ without backpropagation and we reuse the subscript $j$ to $\textbf{p}'$ that is obtained from the sub-image of the $i$-th mixed image corresponding to $j$-th regular image (the subscript $i$ is omitted for convenience). 

% \rv{In terms of consistency regularization, some other forms, \eg, label-prediction pairs, $l_1$ or $l_2$ loss, are also evaluated. We empirically discovered \eq~(\eqref{eq3.5}) achieve the optimal performance, which will be discussed in \sec~{Experiments}.}

\par
Finally, the total training objective of \model-CL is 
\begin{equation}
    \mathcal{L} = \mathcal{L}_{bce} + \mathcal{L}_{cl}~.
\end{equation}

\par
Although it seems that the consistency learning loss is only imposed on the pair of regular images and their sub-images, we claim that it is not simply equal to learning cross-scale knowledge. The better representation of sub-images helps the model discover possibly overlooked objects present in a mixed image, which is beneficial for blending image semantics and improving our \model. 

\section{Experiments}
In this section, we first introduce experimental settings including evaluation metrics and our training details. Next, convergence analyses are given and the proposed methods are verified via plentiful experiments.
% three common MLIC data sets (\ie, Pascal VOC 2007~\cite{everingham2010pascal}, MS-COCO2014~\cite{lin2014microsoft} and Wide Attribute~\cite{li2016human}) on general MLIC and MLIC with missing label tasks, compared to existing state-of-the-art MLIC and Mix-style methods. We also report single-label image classification performance on ImageNet~\cite{russakovsky2015imagenet} and show good transfer learning ability of our \model. 
Finally, we conduct performance analyses to discuss the proposed methods detailedly. 
% \textit{The code will be released at GitHub.}

\subsection{Experimental settings}\label{sec4.1}
\textbf{Evaluation metrics:}
We follow previous works~\cite{chen2019multi,ye2020attention}, adopting mean of Average Precision~(mAP) as the primary metric. Overall precision~(OP), recall~(OR), F1-score~(OF1) and per-class precision~(CP), recall~(CR), F1-score~(CF1) and their top-3 versions are also utilized to evaluate our results. All metric are in \%. For all metrics except mAP, we set a category to be positive if its prediction is larger than 0.5, vice versa. 

\par
\textbf{Training details:}
For the proposed methods and compared methods, we use ResNet-101~\cite{he2016deep} as the base model where the last global average pooling is replaced with global max pooling that is a common setting~\cite{chen2019multi,ye2020attention} to achieve good results for MLIC. We choose SGD as our optimizer with momentum of 0.9 and weight decay of $10^{-4}$. The output dimension of the last fully connected~(fc) layer in ResNet-101 is changed according to the class number of various data sets. We set the learning rate of all layers except the last fc layer to be one-tenth of the given learning rate. During training, we adopt the data augmentation suggested in~\cite{chen2019multi,zhao2021transformer}, \ie, the input image is randomly cropped and resized to 448 × 448 with horizontal flips. We train our model for 80 epochs and decay the learning rate by a factor of 0.1 at 40-th and 60-th epoch. For our \model, we do not seek the optimal grid strategy deliberately and randomly sample $f_{mg}$ from $\mathcal{F}_{mg}=\{f^{1\times 2}, f^{2\times 2}, f^{2\times 3}\}$ with a probability of 0.3 to use a random dropout from 1 to $r\times c-1$ per batch. 
% The number of mixed images is set to a quarter of the regular batch size. 
The cardinality of the mixed set is set to a quarter of the regular batch size, \ie, $|\mathcal{B}^\prime|=\frac{1}{4}|\mathcal{B}|$.
All experiments are implemented based on PyTorch~\cite{paszke2019pytorch}.

\begin{figure}[t]
   \centering  
   \subfigure[Convergence]{
      \includegraphics[width=0.48\linewidth]{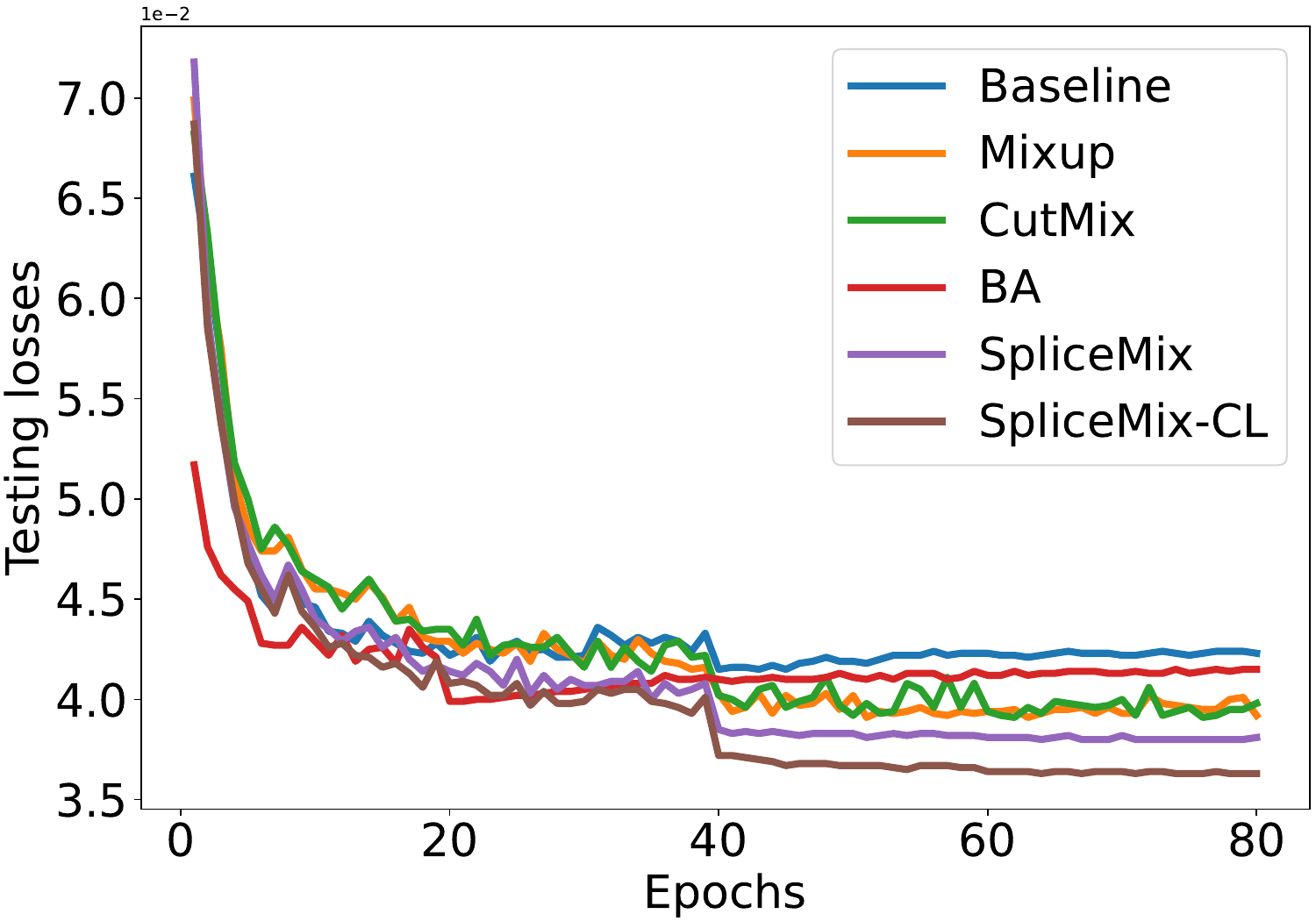}}
   \hfill %\bigskip
   \subfigure[Classification performance]{
      \includegraphics[width=0.48\linewidth]{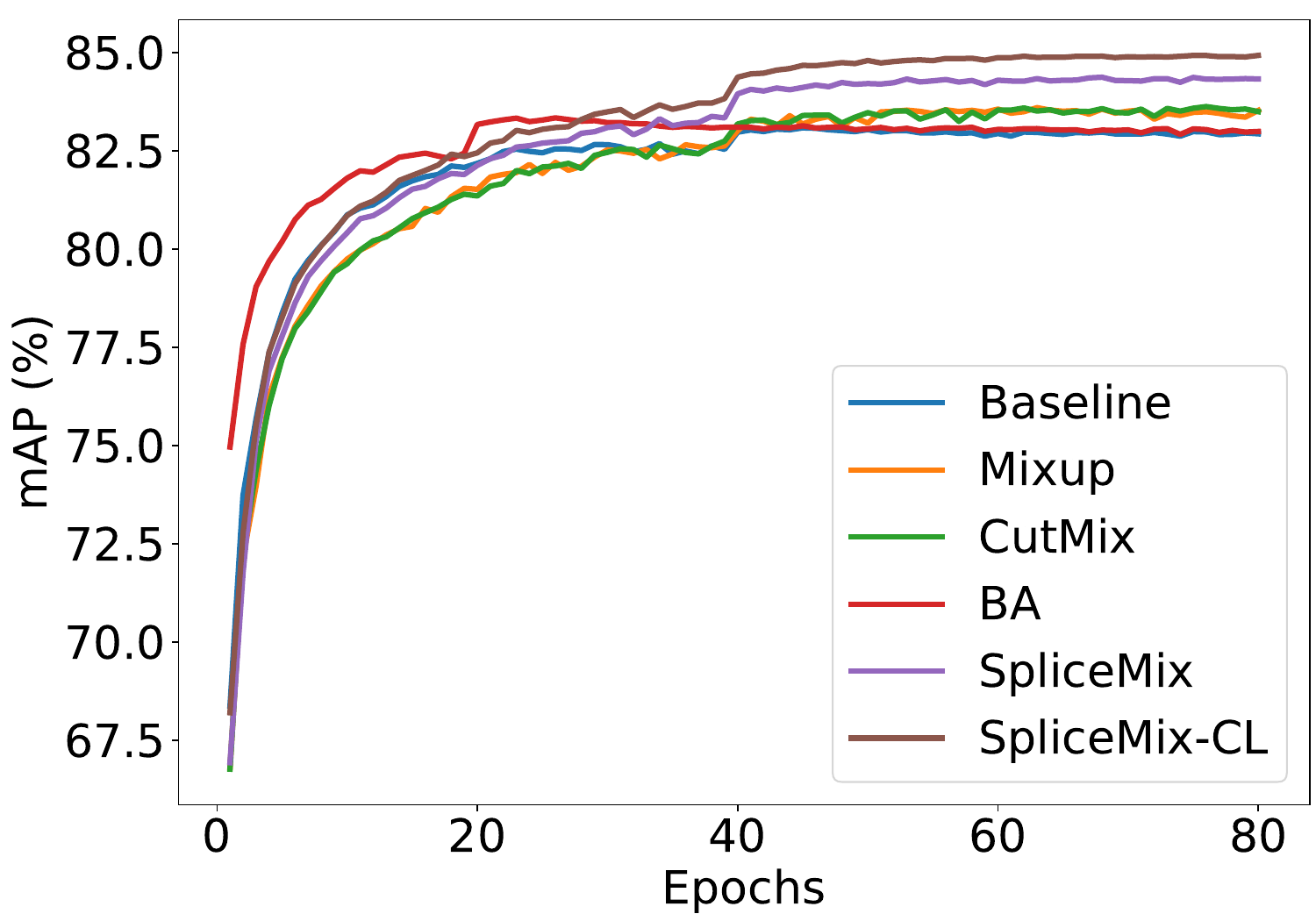}}

   % \vspace{-1.1\baselineskip}
   \caption{Comparisons of convergence and classification performance on MS-COCO~\cite{lin2014microsoft}. We report the results of the total 80 epochs. ResNet-101 is the baseline for all methods. Except for BA, whose learning rate is decayed at 20-th and 40-th epoch, all other methods decay the learning rate at 40-th and 60-th epoch.  
   % ~\cite{he2016deep} with global max pooling (GMP) is utilized as the base model for all other methods. Here we mainly compare our \model\ augmentation method to two popular Mix-style methods, \ie, Mixup~\cite{zhang2018mixup} and CutMix~\cite{yun2019cutmix}. In addition, we also report the results of our \model\ extension~(\model-CL) which cannot be viewed as a augmentation strategy while can show a good potential of our \model.
   }      
   \label{fig4.1}
\end{figure}

\begin{table*}[!th]
\caption{Comparisons with state-of-the-art methods on MS-COCO. The best and second best results (except the methods marked by ``$^*$'') are highlighted in purple and cyan, respectively. The same highlighted way is used for the remaining tables.}  % remove 'highlight' for layout
\label{tb4.3.1}
    \centering
    \small
    % \resizebox{.8\linewidth}{!}{
    \begin{tabular}{c|c|cccccc|cccccc}
        \hline
        \multirow{2}*{Method}         & \multirow{2}*{mAP}  & \multicolumn{6}{c|}{All} & \multicolumn{6}{c}{Top 3 }\\ 
                                      &      & CP   & CR   & CF1  & OP   & OR   & OF1  & CP    & CR   & CF1  & OP   & OR   & OF1  \\ 
         \hline\hline
        % MLGCN~\cite{chen2019multi}           & 83.0 & 85.1 & 72.0 & 78.0 & 85.8 & 75.4 & 80.3 & 89.2  & 64.1 & 74.6 & 90.5 & 66.5 & 76.7 \\
        % SSGRL~\cite{chen2019learning}        & 83.8 & \textcolor{purple}{89.9} & 68.5 & 76.8 & \textcolor{purple}{91.3} & 70.8 & 79.7 & \textcolor{purple}{91.9}  & 62.5 & 72.7 & \textcolor{purple}{93.8} & 64.1 & 76.2 \\
        DSDL~\cite{zhou2021deep}             & 81.7 & 84.1  & 70.4 & 76.7 & 85.1 & 73.9 & 79.1 & 88.1 & 62.9 & 73.4 & 89.6 & 65.3 & 75.6  \\
        CSRA~\cite{zhu2021residual}          & 83.5 & 84.1 & 72.5 & 77.9 & 85.6 & 75.7 & 80.3 & 88.5  & 64.2 & 74.4 & 90.4 & 66.4 & 76.5 \\
        SST~\cite{chen2022sst}               & 84.2 & 86.1 & 72.1 & 78.5 & 87.2 & 75.4 & 80.8 & 89.8  & 64.1 & 74.8 & 91.5 & 66.4 & 76.9 \\
        CCD~\cite{liu2022contextual}         & 84.0 & \textcolor{cyan}{87.2} & 70.9 & 77.3 & \textcolor{purple}{88.8} & 74.6 & 81.1 & 89.7  & 63.9 & 72.9 & 92.0 & 66.5 & 77.2 \\
        IDA-R101~\cite{liu2023causal}        & 84.3 & -    & -    & 78.5 & -    & -    & 81.1 & -     & -    & 73.6 & -    & -    & \textcolor{cyan}{77.3} \\
        \hline
        Baseline                             & 83.0 & 85.0 & 72.0 & 77.9 & 86.9 & 75.2 & 80.3 & 89.0  & 64.1 & 74.5 & 90.7 & 66.3 & 76.6 \\
        Mixup~\cite{zhang2018mixup}          & 83.7 & 84.9 & 72.8 & 78.4 & 86.5 & 75.7 & 80.7 & 89.3  & 64.8 & 75.1 & 91.1 & 66.5 & 76.9 \\
        CutMix~\cite{yun2019cutmix}          & 83.5 & 84.6 & 72.9 & 78.3 & 86.2 & 75.5 & 80.5 & 89.0  & 64.8 & 75.0 & 90.9 & 66.3 & 76.7 \\
        ResizeMix~\cite{qin2020resizemix}    & 83.1 & 86.1 & 71.0 & 77.8 & 88.0 & 73.4 & 80.1 & 90.1  & 64.0 & 74.8 & 91.7 & 65.3 & 76.3 \\
        RecursiveMix~\cite{yang2022recursivemix} &83.4 &86.9 &70.4 &77.8 & \textcolor{purple}{88.8} & 72.6 & 79.9 & \textcolor{cyan}{90.3}  & 63.8 & 74.7 & \textcolor{purple}{92.3} & 64.8 & 76.1 \\
        BA~\cite{hoffer2020augment}          & 83.3 & 84.8 & 72.5 & 78.2 & 85.8 & 75.6 & 80.4 & 89.0  & 64.4 & 74.7 & 90.4 & 66.5 & 76.7 \\
        \hline
        \model                            & \textcolor{cyan}{84.4} & 84.8 & \textcolor{purple}{74.2} & \textcolor{cyan}{79.1} & 86.2 & \textcolor{purple}{77.0} & \textcolor{cyan}{81.3} & 89.2  & \textcolor{purple}{65.6} & \textcolor{cyan}{75.6} & 91.0 & \textcolor{purple}{67.2} & \textcolor{cyan}{77.3} \\
        \modelExt                         & \textcolor{purple}{84.9} & \textcolor{purple}{87.4} & \textcolor{cyan}{73.2} & \textcolor{purple}{79.7} & \textcolor{cyan}{88.2} & \textcolor{cyan}{76.3} & \textcolor{purple}{81.8} & \textcolor{purple}{90.7}  & \textcolor{cyan}{65.1} & \textcolor{purple}{75.8} & \textcolor{cyan}{92.1} & \textcolor{cyan}{67.1} & \textcolor{purple}{77.6} \\
        
        \rowcolor{green! 10}
        \model$^*$
        & 85.8 & 85.1 & 76.6 & 80.6 & 86.4 & 78.7 & 82.3 & 89.6 & 67.3 & 76.9 & 91.4 & 68.2 & 78.1 \\
        \rowcolor{green! 10}
        \modelExt$^*$
        & 86.4 & 85.5 & 76.9 & 81.0 & 87.0 & 79.0 & 82.8 & 89.8 & 67.6 & 77.1 & 91.7 & 68.6 & 78.5 \\
        \hline
    \end{tabular}
    % }
\end{table*}
\subsection{Convergence Analyses}
As we claimed before, our \model\ can lead to faster convergence with lower testing loss. Compared with two popular Mix-style methods (Mixup~\cite{zhang2018mixup} and CutMix~\cite{yun2019cutmix}), BA~\cite{hoffer2020augment} and the baseline ResNet-101, we present the results of convergence and classification performance of testing time in \fig~\ref{fig4.1}. It can be readily seen that our \model\ and \model-CL take the top-2 lowest testing loss and highest mAP. Further, from \fig~\ref{fig4.1}~(a), we can discover that \emph{1)} the baseline and BA fall into over-fitting after 40/20 epochs, 
% since their testing losses turn to rise slightly, 
which results in a little drop of mAP as shown in \fig~\ref{fig4.1}~(b) since its testing loss turns to increase slightly; \emph{2)} both Mixup and CutMix fail to converge steadily, which leads to fluctuant classification performance in \fig~\ref{fig4.1}~(b). As suggested in \cite{yun2019cutmix}, Mixup and CutMix require longer training epochs (twice of the regular training) to achieve good performance. We try our best to follow such epoch setting of \cite{yun2019cutmix} originally leveraged for single-label image classification, while obtaining worse results for MLIC. Hence, we keep the same training setting to our proposed methods for the two Mix-style methods, which is also the optimal setting in our tries. In our proposed \model\ and \modelExt, drawbacks of BA and two Mix-style methods are overcome. In other words, \model\ reduces the risk of over-fitting in the BA and has a steadier and lower convergence bound compared to both Mix-style methods and BA. Furthermore, the proposed \modelExt\ has the lowest testing loss and highest mAP shown in \fig~\ref{fig4.1}, which reveals the excellent extensibility of \model.

\subsection{Multi-label Image Classification~(MLIC)}
% To verify the effectiveness of the proposed \model\ and \modelExt\, we conducts
We evaluate the proposed \model\ and \modelExt\ on two MLIC tasks, \ie, regular MLIC and MLIC with missing labels, where common data sets MS-COCO 2014~\cite{lin2014microsoft} and Pascal VOC 2007~\cite{everingham2010pascal} are adopted. 

\subsubsection{Regular MLIC}
\ \\
\indent\textbf{MS-COCO 2014:} 
Microsoft COCO 2014, short for MS-COCO, is a widely used benchmark for MLIC. There are 82,081 images in the training set and 40,137 images in the validation set with total 80 common objects categories. Since labels of the test set are unavailable, we compare our methods to other methods on the validation set. The learning rate for MS-COCO is set to 0.05 and the batch size is 32. The comparison results are presented in \tb~\ref{tb4.3.1}. It's worth noting that \model\ can achieve comparable performance in ImageNet~\cite{russakovsky2015imagenet} classification. Hence, we also report the results of the two proposed methods adopting ImageNet pre-training based on \model\ and mark them by ``$^*$'', same for \tbs~\ref{tb4.3.2} and \ref{tb4.3.4}. 

\par
To ensure the fairness for compared methods, we make a grid search of the $\alpha$ in Mixup~\cite{zhang2018mixup} and CutMix~\cite{yun2019cutmix} from 0.1 to 1 with a step of 0.2. We find that $\alpha=0.5$ is optimal for the two methods on both MS-COCO and Pascal VOC 2007. For BA, we choose the times of replicating samples (\ie, $M$ in \cite{hoffer2020augment}) from $\{2, 4, 6\}$ and find that replicating samples twice achieves the highest performance.  As shown in \tb~\ref{tb4.3.1}, our two methods have the best performance in most metrics. In compared MLIC methods, CCD~\cite{liu2022contextual} and IDA-R101~\cite{liu2023causal} are designed for removing co-occurred bias which obtain high mAP. With the similar objective, we blend multiple images to generate an unusual scene for existing categories. 
% Beyond the two contextually debiased models, \model also can learn the co-occurrence pattern from regular images 
% The model can learn latent co-occurred relationship of different objects, even these objects are seldom co-occurred. 
Although our \model\ is just a simple augmentation method, it achieves superior performance to two contextually debiased models. Moreover, our \modelExt\ also shows an excellent improvement, compared to \model\ and current advance methods, which suggests the large potential of expanding our \model. From the last two rows of \tb~\ref{tb4.3.1}, we can see that \model-based ImageNet pre-training performs well in the MLIC task.
% More \model\ extensions are worth exploring for MLIC.

\par
\textbf{Pascal VOC 2007:}
Pascal Visual Object Classes Challenge~(VOC 2007) is the most used data set in MLIC. It consists of 20 object categories from 9,963 images. We use the \emph{trainval} set~(5,011 images) for training and the \emph{test} set (4,952 images) for evaluation. The learning rate for Pascal VOC 2007 is set to 0.01 and the batch size is 16. Considering that some current MLIC methods tend to train Pascal VOC 2007 not only with ImageNet pre-training but also with MS-COCO pre-training, where the latter can achieve better performance empirically, we give  results on pre-training from ImageNet and MS-COCO, respectively.
% Pascal VOC 2007~\cite{everingham2010pascal} is widely used multi-label data set, which
% contains 9963 images from 20 common object categories. It is divided into a
% train set, a validation set, and a test set. For fair comparisons, following previous
% works [2,4], we train our model on the trainval set (5011 images) and evaluate
% on the test set (4952 images).

\par
The comparison of our methods and other methods is listed in \tb~\ref{tb4.3.2}. We can see that our \model\ and \modelExt\ outperform all other methods. In compared Mix-style methods, ResizeMix~\cite{qin2020resizemix} and RecursiveMix~\cite{yang2022recursivemix} are highly customized for single-label images that show poor performance (also see \tb~\ref{tb4.3.1}), that is unsuitable for MLIC. Notably, comparing the mAP of four previous Mix-style methods on MS-COCO pre-training that is inferior to the baseline, our \model\ enjoys the gain from MS-COCO hugely. 
% The recognition results are listed in Table~\ref{tb4.3.2}. For fair comparisons, we report mAP and AP of each class on commonly used 448×448 resolution with only ImageNet pretraining. Due to Pascal VOC 2007 is a well-labeled image set, all methods present a small gap in various metrics. Nonetheless, the proposed methods still achieve the best performance.

\begin{table}[t]
\caption{Comparisons with state-of-the-art methods on Pascal VOC 2007. 
% The best and second best results are highlighted in purple and cyan, respectively.
}
\label{tb4.3.2}
    \centering
    \small
    % \resizebox{.8\linewidth}{!}{
    \begin{tabular}{c|cc}
        \hline
        \multirow{2}*{Method}             &      \multicolumn{2}{c}{mAP}    \\ 
                                          &  ImageNet pre.  & MS-COCO pre.  \\
        \hline\hline
        SSGRL~\cite{chen2019learning}     & 93.4           & 95.0         \\
        ML-GCN~\cite{chen2019multi}       & 94.0           & -            \\
        ADDGCN~\cite{ye2020attention}     & -              & 96.0         \\
        CCD~\cite{liu2022contextual}      & -              & 95.8         \\
        ASL~\cite{ridnik2021asymmetric}   & \textcolor{cyan}{94.6}           & 95.8         \\
        \hline
        Baseline~\cite{he2016deep}        & 93.3           & 95.8         \\
        Mixup~\cite{zhang2018mixup}       & 93.8           & 95.6         \\
        CutMix~\cite{yun2019cutmix}       & 93.8           & 95.5         \\
        ResizeMix~\cite{qin2020resizemix} & 93.3           & 95.5         \\
        RecursiveMix~\cite{yang2022recursivemix} & 93.3    & 95.7         \\
        BA~\cite{hoffer2020augment}       & 93.9           & 95.6         \\
        \hline
        \model                              & \textcolor{cyan}{94.6}          & \textcolor{cyan}{96.1}         \\
        \modelExt                           & \textcolor{purple}{94.8}        & \textcolor{purple}{96.3}         \\
        \rowcolor{green! 10}
        \model$^*$                        & 95.3           & 96.5         \\
        \rowcolor{green! 10}
        \modelExt$^*$                     & 95.5           & 96.7         \\
        \hline
    \end{tabular}
    % }
\end{table}

\subsubsection{MLIC with Missing Labels}
\ \\
% \par
\indent  In real scenarios, it is usually difficult to collect all true labels of a multi-label image. An image may be annotated with only partial positive labels, which hinders the performance of MLIC. Multi-label learning with missing/partial-annotated labels has been a hot spot in recent researches~\cite{chen2022structured,zhang2021simple}. In a popular setting~\cite{chen2022structured}, partial positive and negative labels are accurately known and for unknown labels, they can be either positive or negative. In this paper, we follow the previous work \cite{zhang2021simple} and further relax such setting, where no accurate negative labels are required. In other words, only limited positive labels are known in our missing label setting which is more flexible in practice.
% allows lower annotation cost and is more flexible in practice.

\par
We conduct experiments on MS-COCO with missing label setting. Following the setting of \cite{zhang2021simple}, we randomly drop positive labels with different ratios ($80\%$ and $100\%$ but keeping single positive label per image) for constructing the training set. Finally, the remaining labels are $40\%$ of original positive labels and single positive label per image. Keeping the same architecture to previous work~\cite{zhang2021simple}, we adopt ResNet-50 
with global max pooling as our baseline. \tb~\ref{tb4.3.4} lists the comparison results, where methods in the first block is originally reported in \cite{zhang2021simple}, and we implement the remaining methods for comparisons.  It is easy to discover that our methods consistently outperform all other methods. Compared to the baseline, the proposed \model\ shows more robust results as the decrease of left label ratio, which indicates good capacities of our \model\ to handle the problem of missing labels. In addition, the ImageNet pre-training via \model\ helps our two methods boost a lot, which exhibits its advantage in model pre-training. 
% achieve the best performance for the MLIC with missing labels task. Compared to the baseline ResNet-101~(GMP), our method shows more robust results as the decrease of left label ratio. Concretely, when the ratio is $75\%$, \model\ outperforms ResNet-101~(GMP) over $1.1\%$ mAP \textcolor{teal}{similar to the full label situation,} and when the left labels decrease to $40\%$ and single label, the gap of \model\ and ResNet-101~(GMP) increases to $1.4\%$ and $1.8\%$ mAP respectively. The generalizability and robustness of the baseline decline sharply with the increase of missing labels, whereas our \model\ can still maintain certain advantages for MLIC with missing labels, which indicates good capacities of our \model\ to handle missing labels problem from the view of augmentation strategy.

%% tb4.3.3 is not given.
\begin{table}[t]
\caption{Comparisons on MS-COCO with different missing ratio. 
% The best and second best results are highlighted in purple and cyan, respectively.
}
\label{tb4.3.4}
    \centering
    \small
    % \resizebox{.9\columnwidth}{!}{
    \begin{tabular}{c|ccccccccc}
        \hline
        \multirow{2}*{Method}       & \multicolumn{3}{c}{$40\%$ label left} & \multicolumn{3}{c}{single label} \\ 
                                    &         mAP    & CF1  & OF1                  & mAP      & CF1  & OF1  \\ 
                     \hline\hline
        BCE-LS\cite{cole2021multi}         & 73.1    & 44.9 & 41.2                 & 70.5     & 40.9 & 37.3 \\
        WAN\cite{cole2021multi}            & 72.1    & 62.8 & 64.9                 & 70.2     & 58.0 & 58.6 \\
        Focal\cite{lin2017focal}           & 71.7    & 48.7 & 44.0                 & 70.2     & 47.0 & 41.4 \\
        ASL\cite{ridnik2021asymmetric}     & 72.7    & 67.7 & 71.7                 & 71.8     & 44.8 & 37.9 \\
        SPLC\cite{zhang2021simple}         & 75.7    & 67.9 & 73.3                 & 73.2     & 61.6 & 67.4 \\
        \hline
        Baseline\cite{he2016deep}          & 74.4    & 69.3 & 73.5                 & 72.2     & 61.7 & 67.9 \\
        Mixup\cite{zhang2018mixup}         & 75.6    & 70.3 & 74.4                 & 73.6     & 66.9 & 71.6 \\
        CutMix\cite{yun2019cutmix}         & 75.5    & 70.3 & 74.5                 & 73.6     & \textcolor{cyan}{67.3} & 71.6 \\
        ResizeMix~\cite{qin2020resizemix}  & 75.3    & 69.5 & 73.8                 & 73.4     & 67.0 & 71.7 \\
        RecursiveMix~\cite{yang2022recursivemix} &74.9 &69.3 &73.8                 & 73.0     & 66.4 & 70.9 \\
        BA~\cite{hoffer2020augment}        & 74.6    & 69.7 & 73.8                 & 72.5     & 65.7 & 70.0 \\
        \hline                                         
        \model                          & \textcolor{cyan}{76.0}    & \textcolor{cyan}{71.0} & \textcolor{cyan}{75.1}                 & \textcolor{cyan}{74.4}     & \textcolor{purple}{67.5} & \textcolor{cyan}{72.0} \\
        \modelExt                       & \textcolor{purple}{76.9}    & \textcolor{purple}{71.4} & \textcolor{purple}{75.5}                 & \textcolor{purple}{74.7}     & \textcolor{purple}{67.5} & \textcolor{purple}{72.2} \\
        \rowcolor{green! 10}
        \model$^*$                        & 78.6    & 73.5 & 76.5                 & 77.0     & 67.5 & 70.8 \\
        \rowcolor{green! 10}
        \modelExt$^*$                     & 79.6    & 69.7 & 73.8                 & 77.6     & 70.7 & 74.2 \\
        \hline
    \end{tabular}
    % }
\end{table}

\begin{sidewaystable}[!htbp]
\caption{The performance improvement to state-of-the-art MLIC methods using \model. \textcolor{purple}{$\uparrow$} denotes the performance gain and \textcolor{teal}{$\downarrow$} denotes the performance drop. }
% All methods are trained and evaluated in $448\times 448$ image resolution, except ADDGCN, ViT-B and Swin-B. For ADDGCN, it is trained in $448\times 448$ and evaluated in $576\times 576$. For ViT-B and Swin-B, they are trained and evaluated in $384\times 384$ and $224\times 224$, respectively.}
\label{tbC.2}
    \centering
    \small
    \setlength\tabcolsep{1.5pt}
    \begin{tabular}{l|c|cccccc|cccccc}
        \hline
        \multirow{2}*{Method}         & \multirow{2}*{mAP}  & \multicolumn{6}{c|}{All} & \multicolumn{6}{c}{Top 3 }\\ 
                                      &      & CP   & CR   & CF1  & OP   & OR   & OF1  & CP    & CR   & CF1  & OP   & OR   & OF1  \\ 
        \hline\hline
        MLGCN~\cite{chen2019multi}        & 83.0           & 85.1           & 72.0           & 78.0           & 85.8           & 75.4           & 80.3           & 89.2           & 64.1           & 74.6           & 90.5           & 66.5           & 76.7           \\
        \ \ \ \ \ \ \ \ +SpliceMix   & 84.2~\upA{1.2} & 87.1~\upA{2.0} & 72.3~\upA{0.3} & 79.1~\upA{1.1} & 87.8~\upA{2.0} & 75.3~\dnA{0.1} & 81.1~\upA{0.8} & 90.4~\upA{1.2} & 64.8~\upA{0.7} & 75.5~\upA{0.9} & 91.7~\upA{1.2} & 66.6~\upA{0.1} & 77.1~\upA{0.4} \\
                     % & 1.2            & 2.0            & 0.3            & 1.1            & 2.0            & (0.1)          & 0.8            & 1.2            & 0.7            & 0.9            & 1.2            & 0.1            & 0.4            \\
        \hdashline
        SSGRL~\cite{chen2019learning}        & 83.8           & 89.9           & 68.5           & 76.8           & 91.3           & 70.8           & 79.7           & 91.9           & 62.5           & 72.7           & 93.8           & 64.1           & 76.2 \\
        \ \ \ \ \ \ \ \ +SpliceMix   & 84.7~\upA{0.9} & 87.3~\dnA{2.6} & 72.7~\upA{4.2} & 79.3~\upA{2.5} & 87.9~\dnA{3.4} & 75.8~\upA{5.0} & 81.4~\upA{1.7} & 90.5~\dnA{1.4} & 64.6~\upA{2.1} & 75.4~\upA{2.7} & 91.9~\dnA{1.9} & 66.8~\upA{2.7} & 77.4~\upA{1.2} \\
                     % & 0.9            & (2.6)          & 4.2            & 2.5            & (3.4)          & 5.0            & 1.7            & (1.4)          & 2.1            & 2.7            & (1.9)          & 2.7            & 1.2  \\
        \hdashline
        ADDGCN~(576)~\cite{ye2020attention} & 85.2           & 84.7           & 75.9           & 80.1           & 84.9           & 79.4           & 82.0           & 88.8           & 66.2           & 75.8           & 90.3           & 68.5           & 77.9 \\
        \ \ \ \ \ \ \ \ +SpliceMix   & 85.8~\upA{0.6} & 87.6~\upA{2.9} & 74.2~\dnA{1.7} & 80.3~\upA{0.2} & 88.6~\upA{3.7} & 77.1~\dnA{2.3} & 82.5~\upA{0.5} & 91.0~\upA{2.2} & 65.5~\dnA{0.7} & 76.2~\upA{0.4} & 92.5~\upA{2.2} & 67.5~\dnA{1.0} & 78.1~\upA{0.2} \\
                     % & 0.6            & 2.9            & (1.7)          & 0.2            & 3.7            & (2.3)          & 0.5            & 2.2            & (0.7)          & 0.4            & 2.2            & (1.0)          & 0.2  \\
        \hdashline
        TDRG~\cite{zhao2021transformer}         & 84.6           & 86.0           & 73.1           & 79.0           & 86.6           & 76.4           & 81.2           & 89.9           & 64.4           & 75.0           & 91.2           & 67.0           & 77.2 \\
        \ \ \ \ \ \ \ \ +SpliceMix   & 85.1~\upA{0.5} & 86.7~\upA{0.7} & 74.0~\upA{0.9} & 79.8~\upA{0.8} & 87.4~\upA{0.8} & 77.0~\upA{0.6} & 81.9~\upA{0.7} & 90.3~\upA{0.4} & 65.4~\upA{1.0} & 75.8~\upA{0.8} & 91.7~\upA{0.5} & 67.5~\upA{0.5} & 77.7~\upA{0.5} \\
                     % & 0.5            & 0.7            & 0.9            & 0.8            & 0.8            & 0.6            & 0.7            & 0.4            & 1.0            & 0.8            & 0.5            & 0.5            & 0.5  \\
        \hline
        ShuffleNetV2~\cite{ma2018shufflenet} & 69.7           & 80.2           & 51.3           & 62.5           & 85.5           & 58.9           & 69.7           & 82.8           & 47.0           & 60.0           & 88.6           & 54.3           & 67.3 \\
        \ \ \ \ \ \ \ \ +SpliceMix   & 70.7~\upA{1.0} & 77.9~\dnA{2.3} & 56.2~\upA{4.9} & 65.5~\upA{3.0} & 83.6~\dnA{1.9} & 62.6~\upA{3.7} & 71.6~\upA{1.9} & 81.1~\dnA{1.7} & 50.7~\upA{3.7} & 62.4~\upA{2.4} & 88.1~\dnA{0.5} & 56.6~\upA{2.3} & 68.9~\upA{1.6} \\
                     % & 1.0            & (2.3)          & 4.9            & 3.0            & (1.9)          & 3.7            & 1.9            & (1.7)          & 3.7            & 2.4            & (0.5)          & 2.3            & 1.6  \\
        \hdashline
        ResNet-101\footnote{The vanilla ResNet is utilized here. In our main experiments, we replace the last global average pooling of ResNet-101 with global max pooling as our base model for fast convergence and good performance~\cite{chen2019multi} . From this table, we can see our \model\ consistently improve ResNet no matter what the global pooling layer is.}
        ~\cite{he2016deep}   & 80.5           & 83.4           & 68.0           & 74.9           & 86.3           & 72.5           & 78.9           & 87.0           & 60.5           & 71.4           & 90.6           & 64.3           & 75.2 \\
        \ \ \ \ \ \ \ \ +SpliceMix   & 81.9~\upA{1.4} & 79.2~\dnA{4.2} & 74.1~\upA{6.1} & 76.6~\upA{1.7} & 83.2~\dnA{3.1} & 77.1~\upA{4.6} & 80.0~\upA{1.1} & 84.5~\dnA{2.5} & 65.1~\upA{4.6} & 73.5~\upA{2.1} & 89.2~\dnA{1.4} & 66.8~\upA{2.5} & 76.4~\upA{1.2} \\
                     % & 1.4            & (4.2)          & 6.1            & 1.7            & (3.1)          & 4.6            & 1.1            & (2.5)          & 4.6            & 2.1            & (1.4)          & 2.5            & 1.2  \\
        \hdashline
        ViT-B~(384)~\cite{dosovitskiy2020image}& 86.6           & 87.5           & 74.9           & 80.7           & 88.7           & 77.6           & 82.8           & 90.9           & 66.6           & 76.9           & 92.7           & 68.4           & 78.7 \\
        \ \ \ \ \ \ \ \ +SpliceMix   & 87.5~\upA{0.9} & 86.6~\dnA{0.9} & 78.0~\upA{3.1} & 82.1~\upA{1.4} & 87.7~\dnA{1.0} & 80.2~\upA{2.6} & 83.8~\upA{1.0} & 91.4~\upA{0.5} & 68.4~\upA{1.8} & 78.3~\upA{1.4} & 92.4~\dnA{0.3} & 69.5~\upA{1.1} & 79.4~\upA{0.7} \\
                     % & 0.9            & (0.9)          & 3.1            & 1.4            & (1.0)          & 2.6            & 1.0            & 0.5            & 1.8            & 1.4            & (0.3)          & 1.1            & 0.7  \\
        \hdashline
        Swin-B~(224)~\cite{liu2021swin} & 84.9           & 86.4           & 73.7           & 79.6           & 87.5           & 76.3           & 81.5           & 89.9           & 65.8           & 76.0           & 91.6           & 67.7           & 77.8 \\
        \ \ \ \ \ \ \ \ +SpliceMix   & 85.8~\upA{0.9} & 85.4~\dnA{1.0} & 76.2~\upA{2.5} & 80.5~\upA{0.9} & 86.6~\dnA{0.9} & 78.6~\upA{2.3} & 82.4~\upA{0.9} & 90.2~\upA{0.3} & 67.5~\upA{1.7} & 77.2~\upA{1.2} & 91.4~\dnA{0.2} & 68.9~\upA{1.2} & 78.6~\upA{0.8} \\
                     % & 0.9            & (1.0)          & 2.5            & 0.9            & (0.9)          & 2.3            & 0.9            & 0.3            & 1.7            & 1.2            & (0.2)          & 1.2            & 0.8  \\
        \hline        
    \end{tabular}
\end{sidewaystable}

\subsection{Improvement to State-of-the-Art Methods}
The proposed \model\ is a simple and effective augmentation strategy, which is orthogonal to existing MLIC methods and various backbone models. We evaluate our \model\ on the performance improvement to four state-of-the-art MLIC methods and four advanced backbones. The result is listed in \tb~\ref{tbC.2}. In compared methods, Visual Transformer~(ViT)~\cite{dosovitskiy2020image} is fine-tuned from SWAG~\cite{singh2022revisiting} weight and Swin Transformer~(short for Swin)~\cite{liu2021swin} is pre-trained by ImageNet-22k~\cite{ridnik2021imagenet}. All other methods use the ImageNet-1k pre-training.

\par
It can be seen from \tb~\ref{tbC.2}, \model\ consistently improves all methods in primary metrics that are mAP, CF1 and OF1. For state-of-the-art MLIC methods, \model\ boosts them at least 0.5\% and at most 1.2\% mAP. For advanced backbones including a lightweight network~(ShuffleNetV2~\cite{ma2018shufflenet}), a classical CNN~(ResNet-101~\cite{he2016deep}) and two Transformer-based models~(ViT-B and Swin-B), \model\ boosts them at least 0.9\% and at most 1.4\% mAP. 

\par
Multi-label image data sets usually have the issue of positive-negative imbalance~\cite{ridnik2021asymmetric}. A model could predict positive classes with low confidences, leading to high precision and low recall. Our \model\ constructs the label of a mixed image by a union of regular labels, which improves the ratio of positive-negative classes in some degree. Comparing the CP and CR, or OP and OR of four backbones in \tb~\ref{tbC.2}, we can discover that CP and OP have larger values than CR and OR due to too low confidences of positive classes. In our \model, such circumstance is mitigated. \model\ helps these backbones increase the CR and OR with an acceptable drop of CP and OP. As a result, \model\ improves the CF1 and OF1 significantly.

\subsection{Small Object Recognition}
A multi-label image usually contains multiple objects whose scales are fickle. Small objects 
could be ignored due to low resolution~\cite{zhao2021transformer} or global pooling operation~\cite{liu2021query2label}, resulting in poor recognition performance. In \model, we deal with small object recognition problem via exploiting information learned from images with different scales in the same batch efficiently. Besides, we also introduce a consistency learning-based design to bridge the gap of objects between regular images and mixed images to further improve \model. 

\begin{figure}[ht]
   \centering  
   \subfigure[\model]{
      \includegraphics[width=1\linewidth]{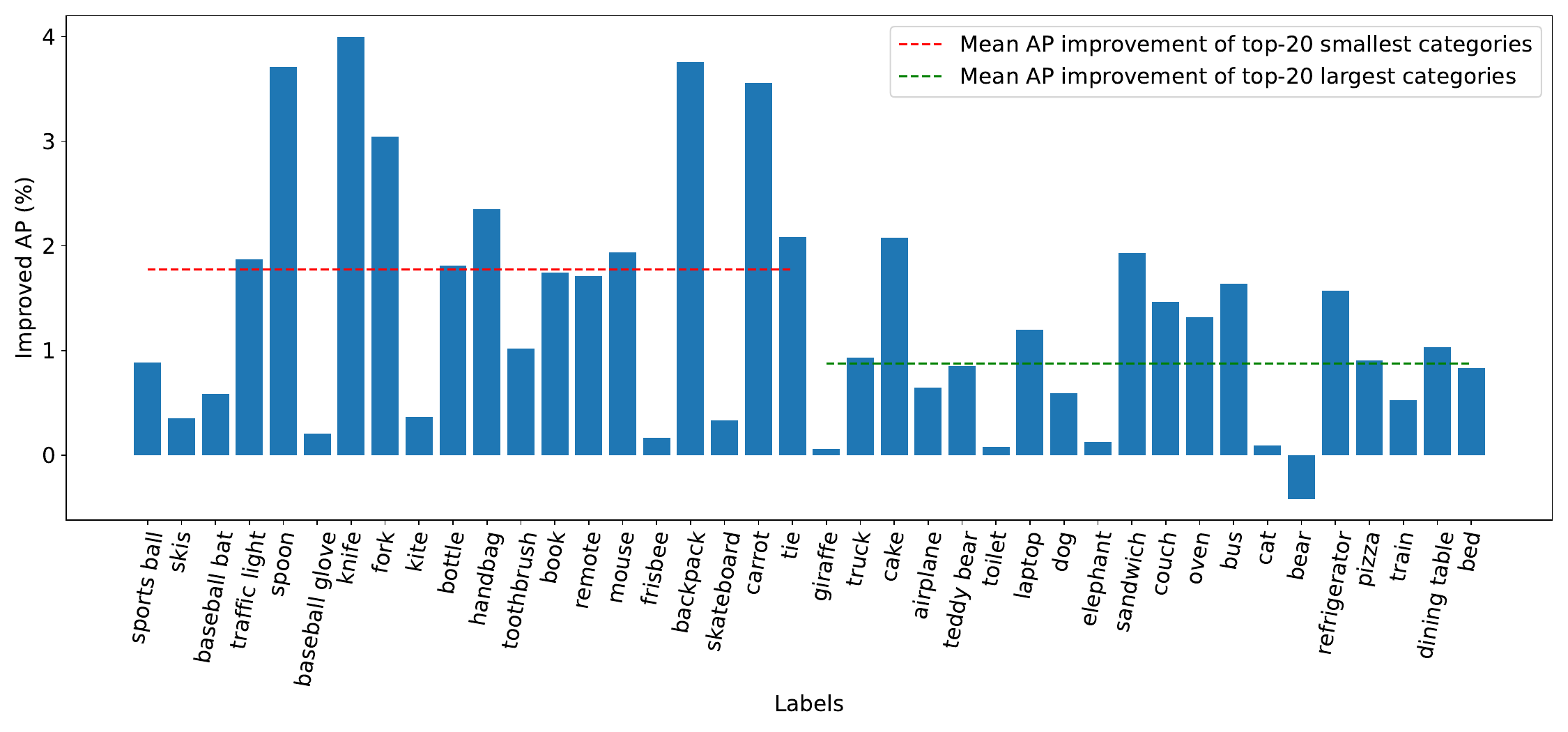}}  % 1.77 0.87
   \hfill %\bigskip
   \subfigure[\modelExt]{
      \includegraphics[width=1\linewidth]{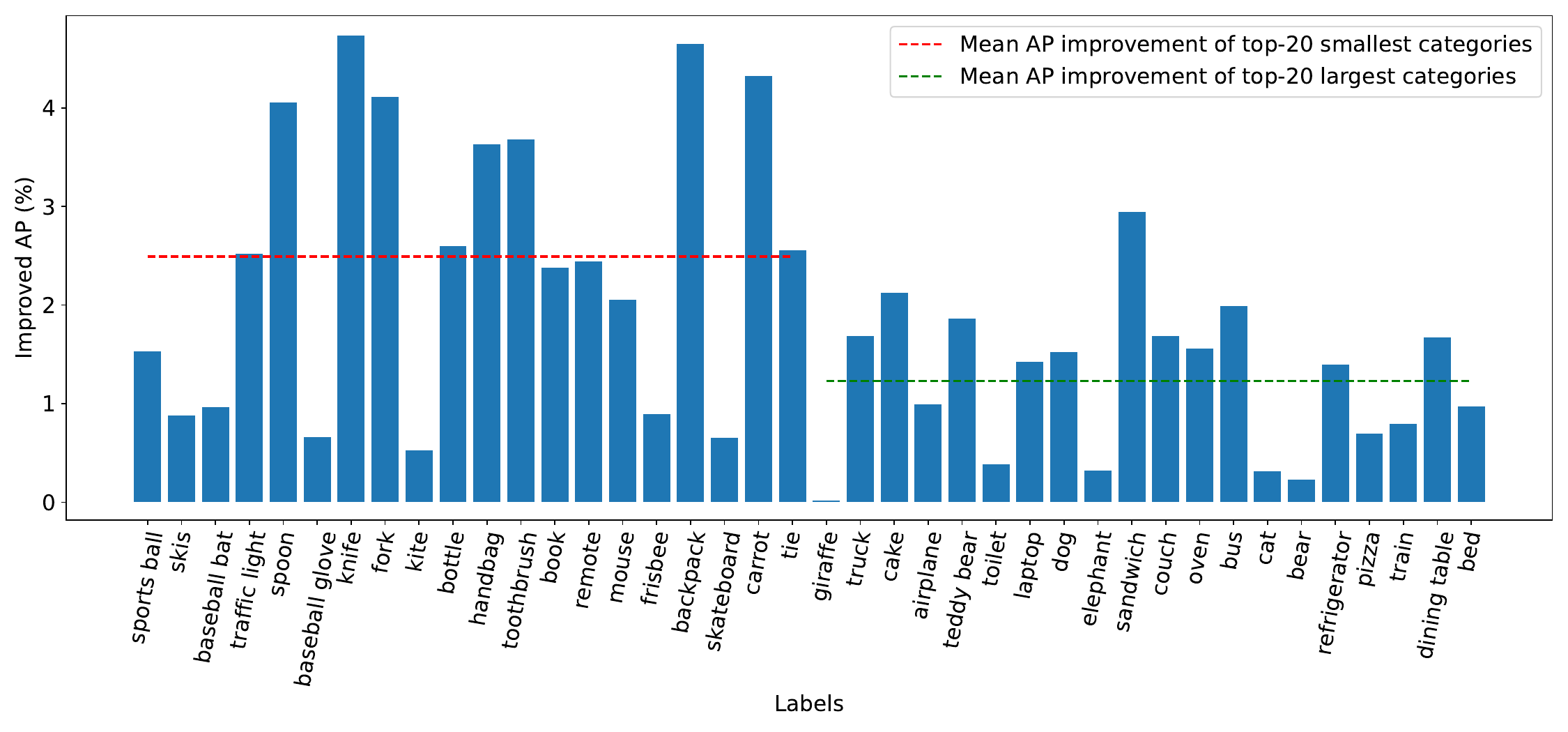}}  % 2.5 1.23

   % \vspace{-1.1\baselineskip}
   \caption{AP improvement of our methods boosting the baseline. For clear presentation, the top-20 smallest and largest categories are selected from left to right.
   }      
   \label{figB.2}
\end{figure}

We conduct experiments on MS-COCO to evaluate our methods in boosting small object recognition performance. In fact, different from the object detection task, it is impossible to recognize all objects in a multi-label image due to the lack of bounding boxes, although small objects always impair the performance of MLIC. We calculate the average object size of each categories by the bounding box offered in the MS-COCO data set. And then, we sort all categories from small to large according to the average object size. The AP improvement of our methods compared to the baseline (ResNet-101) is illustrated in \fig~\ref{figB.2}. The proposed \model\ and \modelExt\ boost the small object recognition performance remarkably. The mean AP improvement of the top-20 smallest and largest categories in \model\ is 1.8\% and 0.9\%, respectively. \model\ boosts the AP of small objects more than twice that of large objects. Comparing \figs~\ref{figB.2}~(a) and (b), \modelExt\ further improves \model\ in almost all selected categories, which indicates the effectiveness of our consistency learning-based design and the flexible extensibility of \model.

\par 
Note that the improvement of small object recognition is not only due to cross-scale training in our \model, but also owing to semantic blending learning. Small objects tend to have poor correlation to other objects because they could be neglected easily. Our \model\ enhances the label dependency and reduces the co-occurred bias that can contribute to identify small objects both in their usual and unusual scenes.

\subsection{Inference with Low Resolution Images}\label{sec4.5}
In practical applications, the quality of query images may be not guaranteed. A model could be trained with high resolution images, while inferring images under a low resolution circumstance. The model performance will be hindered severely, due to the large resolution discrepancy between training and inferring~\cite{touvron2019fixing}. Our \model\ trains regular images and their downsampled versions simultaneously, which is beneficial to mitigate the above issue.

\begin{figure}[t]
   \centering  
   \includegraphics[width=1\linewidth]{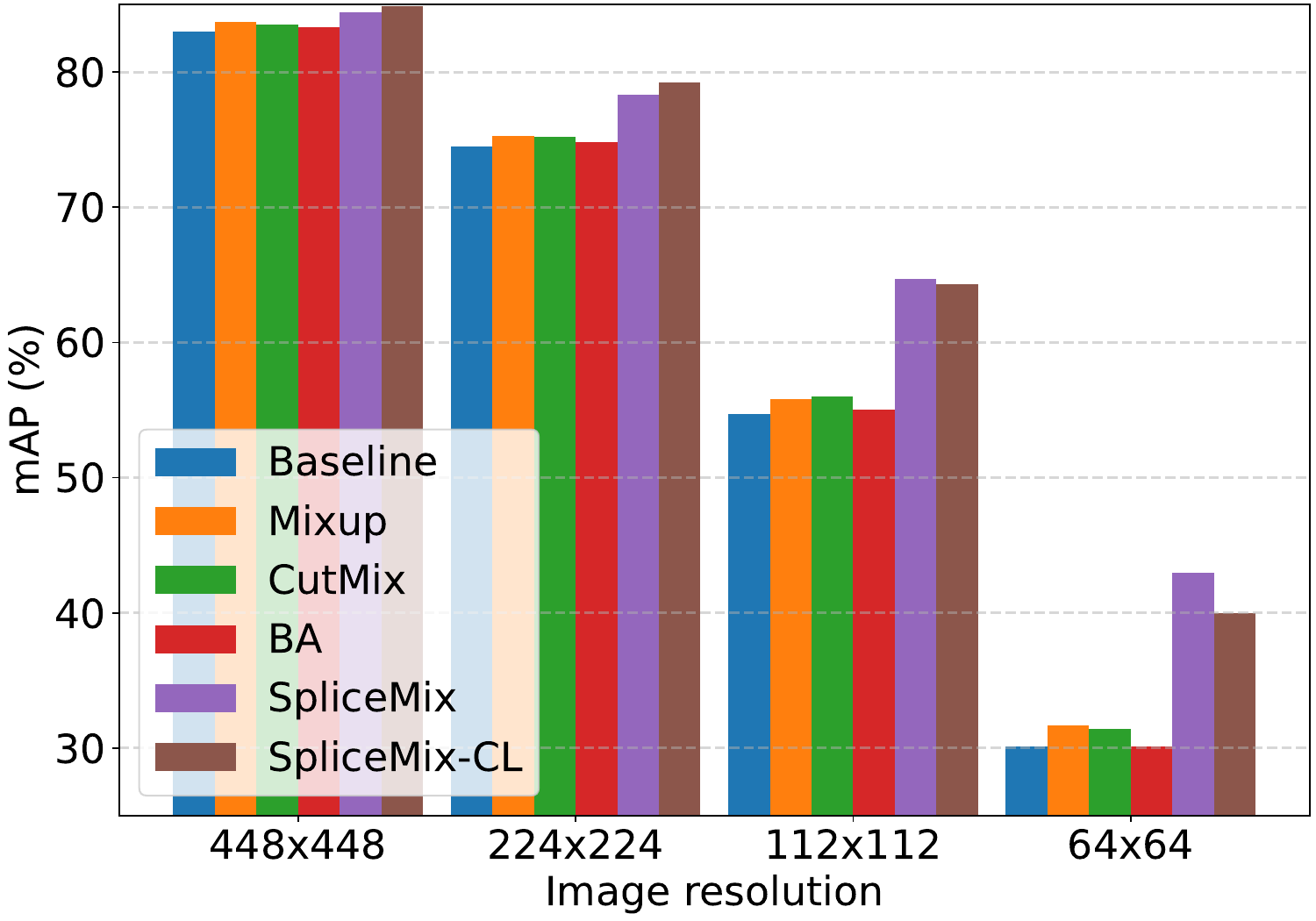}

   \caption{Comparisons on various image resolutions in inferring phase. All methods are trained with $448\times 448$ image resolution and inferred with the resolutions of $448\times 448$, $224\times 224$, $112\times 112$, and $64\times 64$, respectively.}
   % The compared results in $448\times 448$ are same to \tb~\ref{tb4.3.1}.}      
   \label{figC.1}
\end{figure}

\fig~\ref{figC.1} compares the inference performance with various image resolutions on MS-COCO. We can see that low resolutions degrade the performance of all methods. The mAP of four compared methods drops rapidly as the inferring resolution declines. However, our \model\ and \modelExt\ present good robustness to low resolution images. For instance, the mAP of \model\ surpasses compared methods over at least 10\% in the ultra-low $64\times 64$ resolution. Besides, we may discover that \modelExt\ outperforms \model\ in the resolutions of $448\times 448$ and $224\times 224$, while losing its superiority in $112\times 112$ and $64\times 64$. In the default setting of \model~(see \sec~\ref{sec4.1}), at least one of the height and width of a downsampled image is 224. The consistency learning between regular images and their downsampled images may focus on bridging the gap between $448\times 448$ and $224\times 224$, resulting in sub-optimal performance to lower resolutions. 

\subsection{Performance Analyses}
\subsubsection{Label Dependency and Co-occurred Bias}
\ \\
\indent
In the multi-label image classification task, it is challenging to balance the label dependency and co-occurred bias. Capturing label dependency is beneficial for a model to recognize co-occurred categories in their usual context. However, it may hamper the model generalizability of recognizing categories occurring in their unusual context. The proposed \model\ learns label dependency from the regular batch and reduces co-occurred bias via semantic blending learning in the mixed set. The ameliorated label dependency can be built from the trade-off between learning label dependency and reducing co-occurred bias.

\begin{figure}[!ht]
   \centering  
   \subfigure[Baseline]{
        \begin{minipage}{1\linewidth}
             \centering
             \includegraphics[width=.997\linewidth]{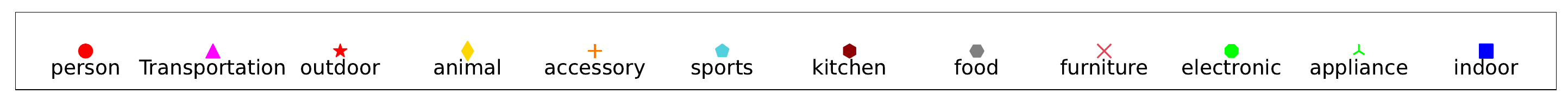}
             \includegraphics[width=1\linewidth]{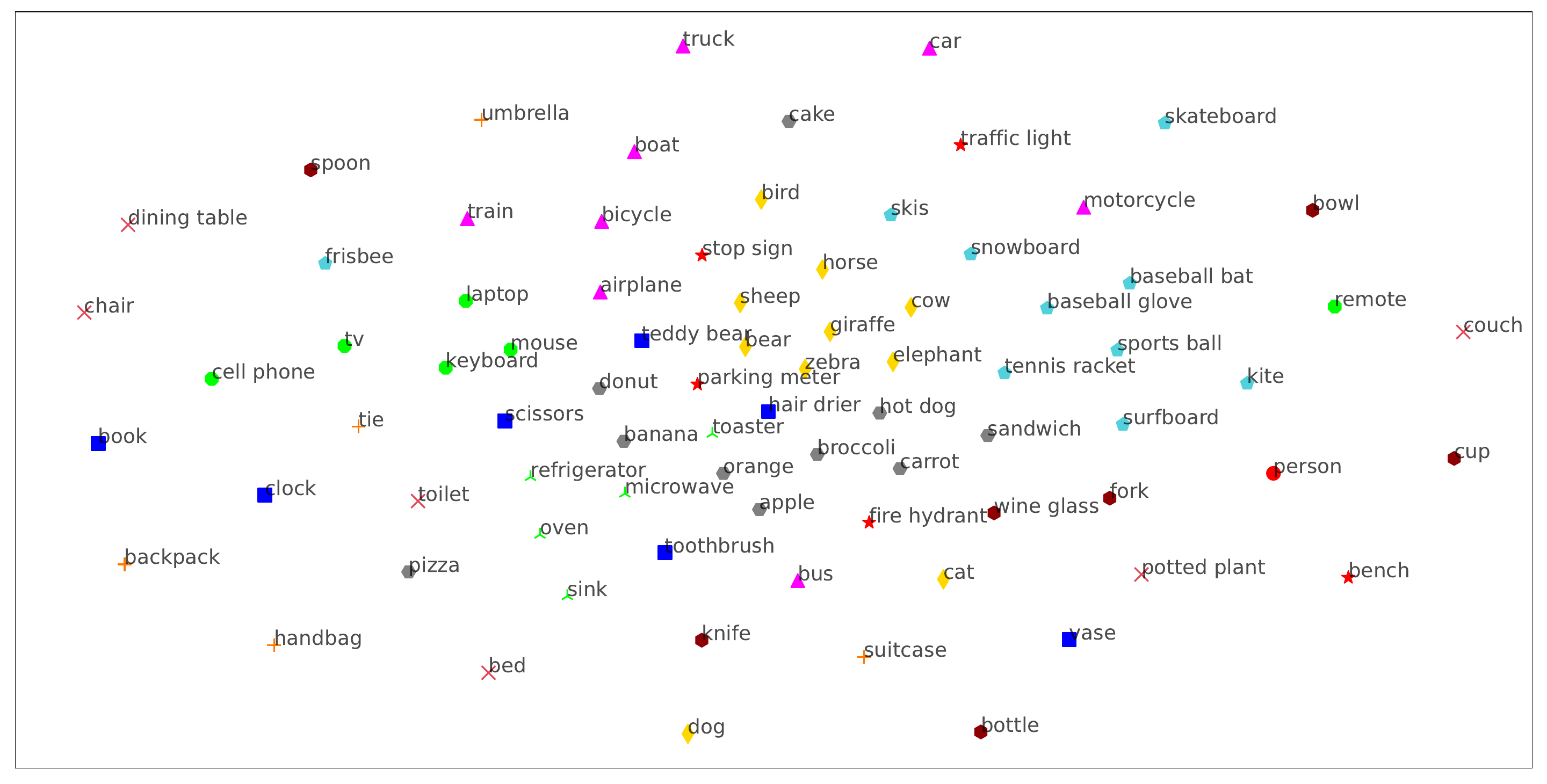}
        \end{minipage}
      }
   \hfill %\bigskip
   \subfigure[\model]{
      \includegraphics[width=1\linewidth]{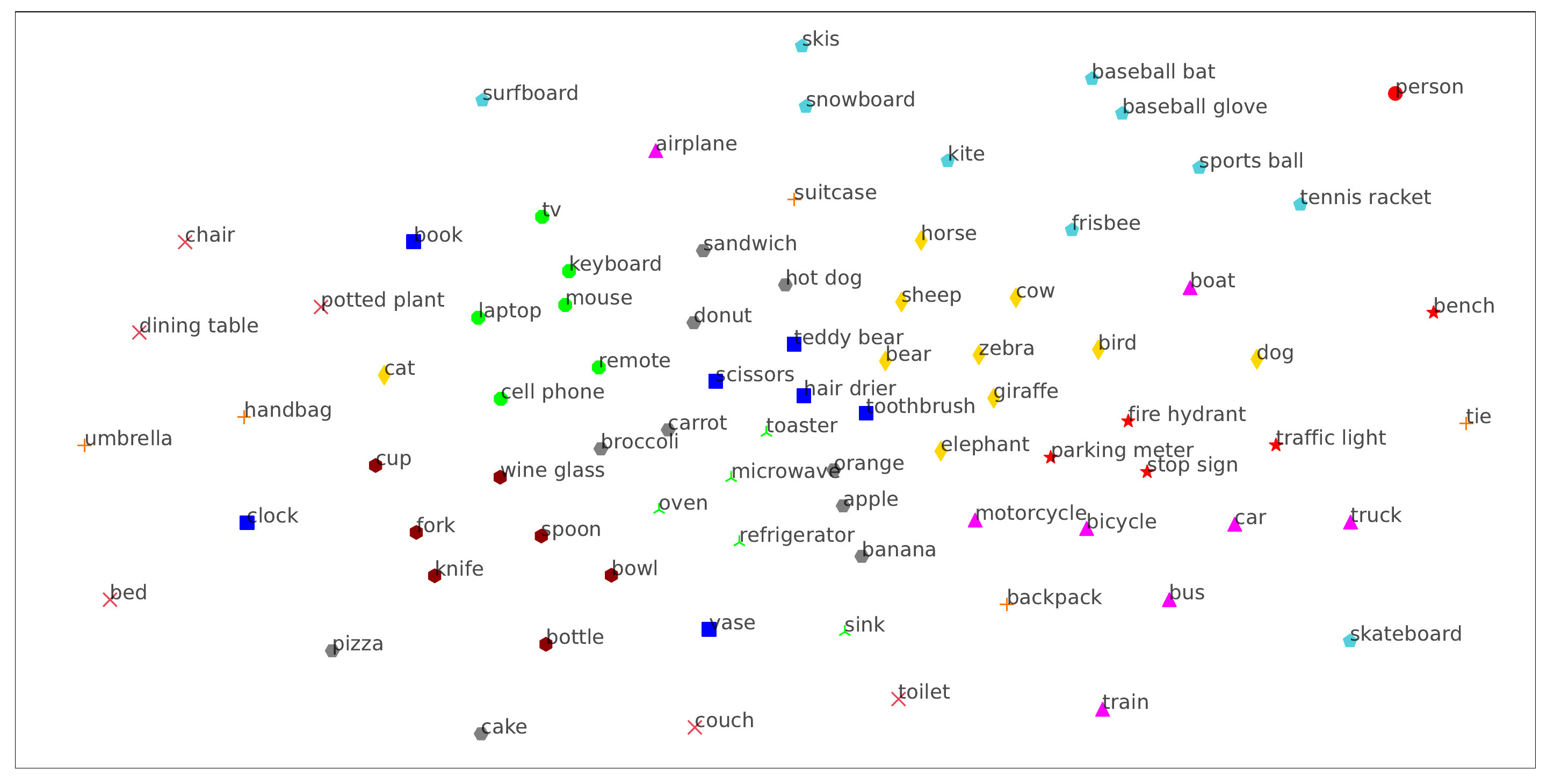}}  % 2.5 1.23

   \caption{T-SNE Visualization on the classifier weight of the baseline and \model. Categories in the same shape and color belong to the same super set offered by MS-COCO.}      
   \label{figB.1.1}
\end{figure}

To verify this, we first give the result of label dependency learning in our \model. The T-SNE visualization on the classifier weight pre-trained on MS-COCO~\cite{lin2014microsoft} is presented in \fig~\ref{figB.1.1}. Compared to the baseline~(ResNet-101~\cite{he2016deep}), \model\ can capture better label correlation. For example, the categories ``cup'' and ``bowl'' lose their co-occurrence relationships in the baseline shown in \fig~\ref{figB.1.1}~(a). However, in our \model, the two categories are near their correlated categories, which accurately reflects category relationships in their super set ``kitchen''. Moreover, \model\ also builds improved label dependencies for other super sets, such as ``electronic'', ``animal'' and ``outdoor''. 

\begin{figure}[t]
   \centering  
   % \hspace*{-2cm}
   \includegraphics[width=1\linewidth]{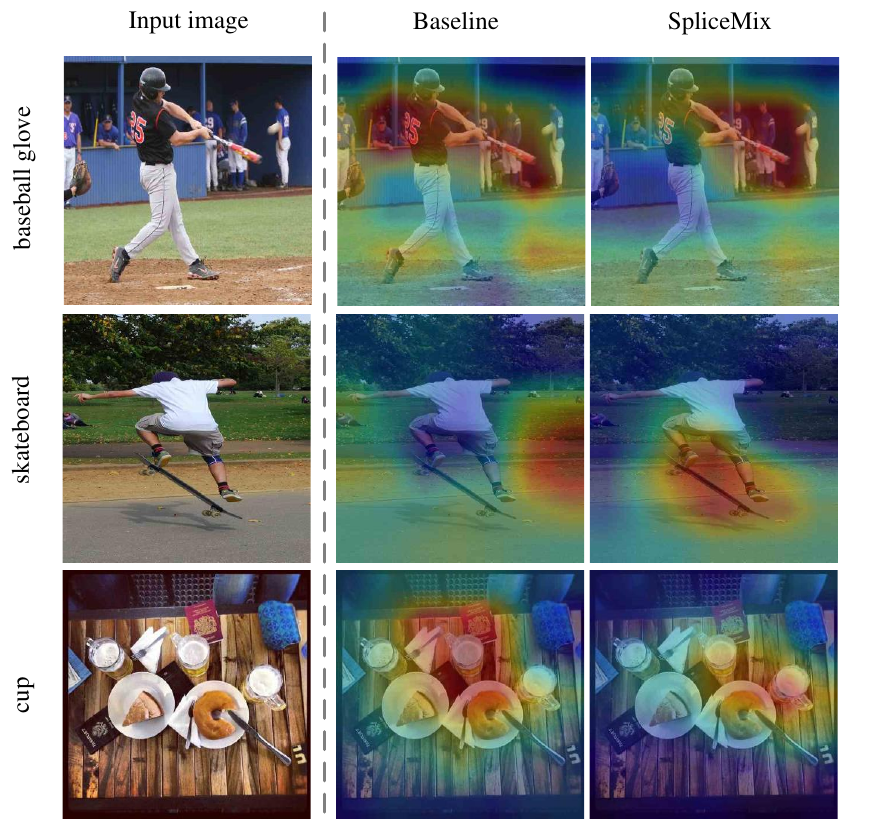}
   \caption{CAM visualization of highly biased categories. The biased category pairs (baseball\ glove, person), (skateboard, person) and (cup, dining\ table) are in the 20 most biased category pairs.}      
   \label{figB.1.2}
\end{figure}

\par
For studying co-occurred bias, we follow the previous work~\cite{singh2020don} and conduct experiments on COCO-Stuff~\cite{caesar2018coco} that is an augmented version of MS-COCO 2017 with pixel-wise annotations for 91 stuff classes. We run our \model\ based on a re-implementation
% \footnote{\href{https://github.com/princetonvisualai/ContextualBias}{https://github.com/princetonvisualai/ContextualBias}} 
of \cite{singh2020don}. Before evaluation, the 20 most biased category pairs are determined. For each biased category pair $(b, c)$, the test set can be divided into three sets: co-occurred images that contain both $b$ and $c$, exclusive images that contain $b$ but not $c$, and other images that do not contain $b$. Then, two test distributions can be constructed: 1) the ``exclusive'' distribution containing exclusive and other images and 2) the ``co-occur'' distribution containing co-occurred and other images. More details can be found in \cite{singh2020don,kim2021re}. The classification result is reported in \tb~\ref{tbB.1}. 
\begin{table}[t]
\caption{Performance comparison on exclusive distribution and co-occurred distribution. The mAP metric is reported.}
% ``Exclusive+Co-occur'' denotes the sum of mAP of ``Exclusive'' and ``Co-occur''. }
\label{tbB.1}
    \centering
    \small
    \begin{tabular}{c|ccc}
        \hline
        Method    & Exclusive & Co-occur & Exclusive+Co-occur \\
        \hline\hline
        Baseline  & 21.6      & 65.5     & 87.1               \\
        SpliceMix & \textcolor{purple}{23.7}      & \textcolor{purple}{66.4}     & \textcolor{purple}{90.1}              \\
        \hline
    \end{tabular}
\end{table}
As we can see, \model\ improves the baseline~(ResNet-50) by 2.1\% mAP in exclusive distribution, which means our \model\ has better performance of recognizing categories when these categories occur in their unusual context. The co-occurred bias is mitigated by \model. In addition, for co-occurred categories, \model\ also outperforms the baseline by 0.9\% mAP, which indicates ameliorated label dependency is learned in \model. A good trade-off between learning label dependency and reducing co-occurred bias can be sought. Furthermore, we compare the class activation map~(CAM)~\cite{zhou2016learning} of the baseline and \model\ to present the issue existing in highly biased category pairs. As shown in \fig~\ref{figB.1.2}, for the categories ``baseball glove'' and ``cup'', the baseline locates more areas on the objective co-occurred categories, \ie, ``person'' and ``dining table'', respectively. Only a little information of the query category can be learned which may be less discriminative for recognizing this category in its rare context. Oppositely, our \model\ locates ``baseball glove'' and ``cup'' intensively, which means that less co-occurred bias is induced. Let us see the second row of \fig~\ref{figB.1.2}, the baseline fails to discover the object, which could be because the baseline can not identify the easy category ``person'', resulting in that its correlated object (\ie, ``skateboard'') is ignored. In our \model, such strong co-occurred relationship is not required. Hence, \model\ can recognize and locate ``skateboard'' effectively.

\subsubsection{Ablation Studies}
\ \\
\indent
We study the effectiveness of the proposed methods from four aspects, \ie, cross-scale learning, semantic blending learning, batch splice and consistency learning, where the batch splice denotes the splice of the regular batch and mixed set~(see \fig\ref{fig3.1}). Considering that the batch splice and consistency learning is not orthogonal to cross-scale learning and semantic blending learning, we will discuss them among the latter. The experiments are conducted on MS-COCO with a $2\times 2$ grid strategy of \model\ and the results are listed in \tb~\ref{tb4.3.7} where the denotations $f_{ds}$ and $f_{mg}$ account for cross-scale learning and semantic blending learning, respectively~(see \eq~\eqref{eq3.1}). $\cup$ denotes batch splice and $L_{cl}$ denotes consistency learning (see \eq~\eqref{eq3.5}).. 
% As shown in \fig~\ref{fig3.1}, both regular images and their downsampled versions are fed into feature extractor simultaneously. The object information from different scales are learnt in the same \model ed batch, improving recognition of small objects~\textcolor{teal}{(see supplements)}. Each image from the mixed set is a combination of multiple downsampled regular images, blending image semantics and reducing scene bias. Owing to the splice of regular batch and mixed set, we can build the consistency learning between mixed images and regular images for boosting the performance further. We conduct experiments on MS-COCO with $2\times 2$ grid setting and list the results in Table~\ref{tb4.3.7}. Considering that the batch splice and consistency learning is not orthogonal to cross-scale learning and semantic blending learning, we will discuss them among the latter. 

\textbf{Cross-scale learning:}
For removing the component of semantic blending, we adopt the setting of grids with dropout, \ie, $2\times 2-3$. Then, the mixed image is degraded to a downsampled image merely, 
% Concretely, we randomly drop 3 sub-images in a mixed image with $2\times 2$ grid, and then such mixed image degrades to a downsampled version of the left regular image, 
which is corresponding to ($f_{ds}+\cup$) in \tb~\ref{tb4.3.7}. The consistency learning can be built from regular images and their downsampled images, corresponding to ($f_{ds}+\cup+L_{cl}$). To reach only $f_{ds}$, we downsample regular images to half of their height and width, and then train downsampled images after training regular batch. From \tb~\ref{tb4.3.7}, we can see that the batch splice improves the performance of cross-scale learning, which indicates the cross-scale information from the same batch is better than that from different batch. Further, with the consistency learning, the performance of cross-scale learning is boosted a lot. In our \modelExt, the knowledge of fine regular images is leveraged to guide the model to learn better representation of coarse mixed images. In this part of only cross-scale learning, the gain from consistency learning is obvious, and it will be more with the joint of semantic blending learning. Besides, owing to cross-scale learning, \model\ has an advantage of recognizing small objects, even when the query image is low-resolution (see \sec~\ref{sec4.5}).

\textbf{Semantic blending learning:}
% For removing the component of cross-scale, we obtain mixed images via making a grid of regular images without downsampling. To reach only $f_{mg}$, we train these mixed images after regular batch. Since mixed images have twice height and width than regular ones, we cannot splice the mixed set and regular batch. We forward the mixed set and regular batch respectively while accumulating their gradients for backpropagation, which is corresponding to ($f_{mg}$+$\cup$). After that, we can add consistency learning loss to the part of forwarding the mixed set via exploiting the prediction logits of regular batch to supervise those of mixed set, corresponding to ($f_{mg}$+$\cup$+$L_{cl}$).
Semantic blending in our \model\ is implemented by making a grid of several downsampled regular images. The object information is lossless except the resolution, which better contributes to producing an unusual scene for present objects than current Mix-style methods~\cite{zhang2018mixup,yun2019cutmix}. Evaluating the semantic blending learning separately is spiny. Therefore, we evaluate its performance by incremental comparisons. Comparing ($f_{ds}+\cup$) to ($f_{ds}+f_{mg}+\cup$) in \tb~\ref{tb4.3.7}, we can find that with semantic blending, the mAP is improved by 0.7\% that suggests the effectiveness of semantic blending learning to mitigate the co-occurred bias. We may also notice that ($f_{ds}+f_{mg}$) without batch splice harms the performance a little compared to the baseline, from which we can realize the importance of batch splice.

\begin{table}[t]
\caption{Ablation study for different components of \model\ and \model-CL. The first row is the results of the baseline and the last two rows are the results of \model\ and \modelExt, respectively.}
\label{tb4.3.7}
    \centering
    \small
    % \resizebox{.7\linewidth}{!}{
    \begin{tabular}{ccccccc}
        \hline
         $f_{ds}$   & $f_{mg}$   & $\cup$     & $L_{cl}$   & mAP  & CF1  & OF1   \\
        \hline\hline
                     &           &            &            & 83.0 & 77.9 & 80.3  \\
        \hdashline
        \checkmark &             &            &            & 83.2 & 77.3 & 80.0  \\
        \checkmark &             & \checkmark &            & 83.5 & 78.4 & 80.6  \\
        \checkmark &             & \checkmark & \checkmark & 84.1 & 78.5 & 80.7  \\
        % \hdashline
        \checkmark & \checkmark  &            &            & 82.9 & 77.8 & 80.2  \\
        \hdashline
        \checkmark & \checkmark & \checkmark &            & 84.2 & 78.7 & 80.9  \\
        \checkmark & \checkmark & \checkmark & \checkmark & 84.7 & 79.7 & 81.7  \\
        \hline
    \end{tabular}
    % }
\end{table}

\subsubsection{Ways of Batch Splice}
\ \\
\indent
The batch splice in our \model\ is a combination of the mixed set and regular batch, where the final \model ed batch size is increased a little due to the introduction of the mixed set. Another way to batch splice is combining the mixed set and regular images that do not attend to mixing. In other words, there is no repeated image in such final batch whose size is less than the regular batch. We conduct experiments on MS-COCO to compare the performance of the batch splice used in our \model\ and the batch splice without repeated images. In our \model, the regular batch size is 32 and the cardinality of the mixed set is 8. To reach the same final batch size to \model, for the compared method, we choose the number of regular images and mixed images from the pairs of $\{(32, 8), (20, 20), (0, 40)\}$, where the pair of $(0, 40)$ only uses the mixed images for training. A $2\times 2$ grid strategy is adopted for all methods.

\tb~\ref{tbC.3.1} reports the results of the two ways of batch splice. When only mixed images are used (\ie, $(0,40)$ in \tb~\ref{tbC.3.1}) for training, the model performance degrades a lot. In the compared method, the mAP increases gradually with the decreased number of mixed images, which indicates the splice with regular images contributes to boosting performance. However, the performance boost of the compared methods is limited. With the same number of regular images and mixed images to our \model, the compared method achieves 83.7\% mAP that is less than \model. As shown in \tb~\ref{tbC.3.1}, the batch splice used in our \model\ is superior to the compared method.

% To be honest, the compared method generates mixed images from regular images which will no longer contribute to training, accelerating training.

\begin{table}[t]
\caption{Comparisons of batch splice used in \model~(Ours) and without repeated images. $(\cdot,\cdot)$ denotes the number of regular images (the former) and mixed images (the latter), respectively, same for \tb~\ref{tbC.4.1}}
\label{tbC.3.1}
    \centering
    \small
    \begin{tabular}{c|cccc}
        \hline
        Method & Ours & $(32, 8)$ & $(20, 20)$ & $(0, 40)$ \\
        \hline\hline
        mAP    & \textcolor{purple}{84.2} & 83.7    & 82.6     & 81.0    \\
        \hline
    \end{tabular}
\end{table}

\subsubsection{Batch Splice for Previous Mix-style Methods}
\label{secC.4}
\ \\
\indent
As we claimed in this paper, our \model\ is designed for MLIC according to their characters. The mixed images in \model\ preserve unbroken objects and blend the image semantics for alleviating co-occurred bias, which is more suitable for MLIC than previous Mix-style methods. To demonstrate this, we replace the the part of our mixing with two popular Mix-style methods~(\ie, Mixup~\cite{zhang2018mixup} and CutMix~\cite{yun2019cutmix}). We compare our \model\ with the splice of regular batch and mixed images from Mixup and CutMix, respectively. In this experiment, the regular batch size is 32 for MS-COCO~\cite{lin2014microsoft}. We consider two settings for compared methods, which are the cardinality of the mixed set same to regular batch size and same to that used in \model. For the former setting, we set the regular batch size to 20 and the final batch size for training will be 40 to reach the same final batch size to other methods. In our \model, the cardinality of the mixed set is 8. 
% A $2\times 2$ grid strategy is adopted for all methods.

The results of splice with Mixup and CutMix are listed in \tb~\ref{tbC.4.1}. As we can see, 
mixed images generated by Mixup cannot work well with the joint of regular images. Mixup linearly combines two images, which may be too complex for MLIC. The large difference between regular images and mixed images from Mixup leads to poor learning ability in a spliced batch. On the other hand, the splice with CutMix achieves similar mAP to vanilla CutMix, no matter how many mixed images are used. Compared to the splice with Mixup and CutMix, our \model\ is the optimal way for batch splice, since mixed images from \model\ are suitable for MLIC training and perform well with the joint of regular images.

\begin{table}[t]
\caption{Comparisons of batch splice with different Mix-style methods.}
% $(\cdot,\cdot)$ denotes the number of regular images (the former) and mixed images (the latter), respectively.}
\label{tbC.4.1}
    \centering
    \small
    \begin{tabular}{c|cc}
        \hline
        \multirow{2}*{Method}             & \multicolumn{2}{c}{mAP}   \\ 
                                          & $(32, 8)$     & $(20, 20)$  \\
        \hline\hline
        Mixup~\cite{zhang2018mixup}       & 83.0          & 82.2      \\
        CutMix~\cite{yun2019cutmix}       & 83.6          & 83.6      \\
        \model                            & \textcolor{purple}{84.4}          & -         \\
        \hline
    \end{tabular}
\end{table}

\subsubsection{Performance on Various Grids}
\begin{figure}[t]
   \centering  
   \includegraphics[width=1\linewidth]{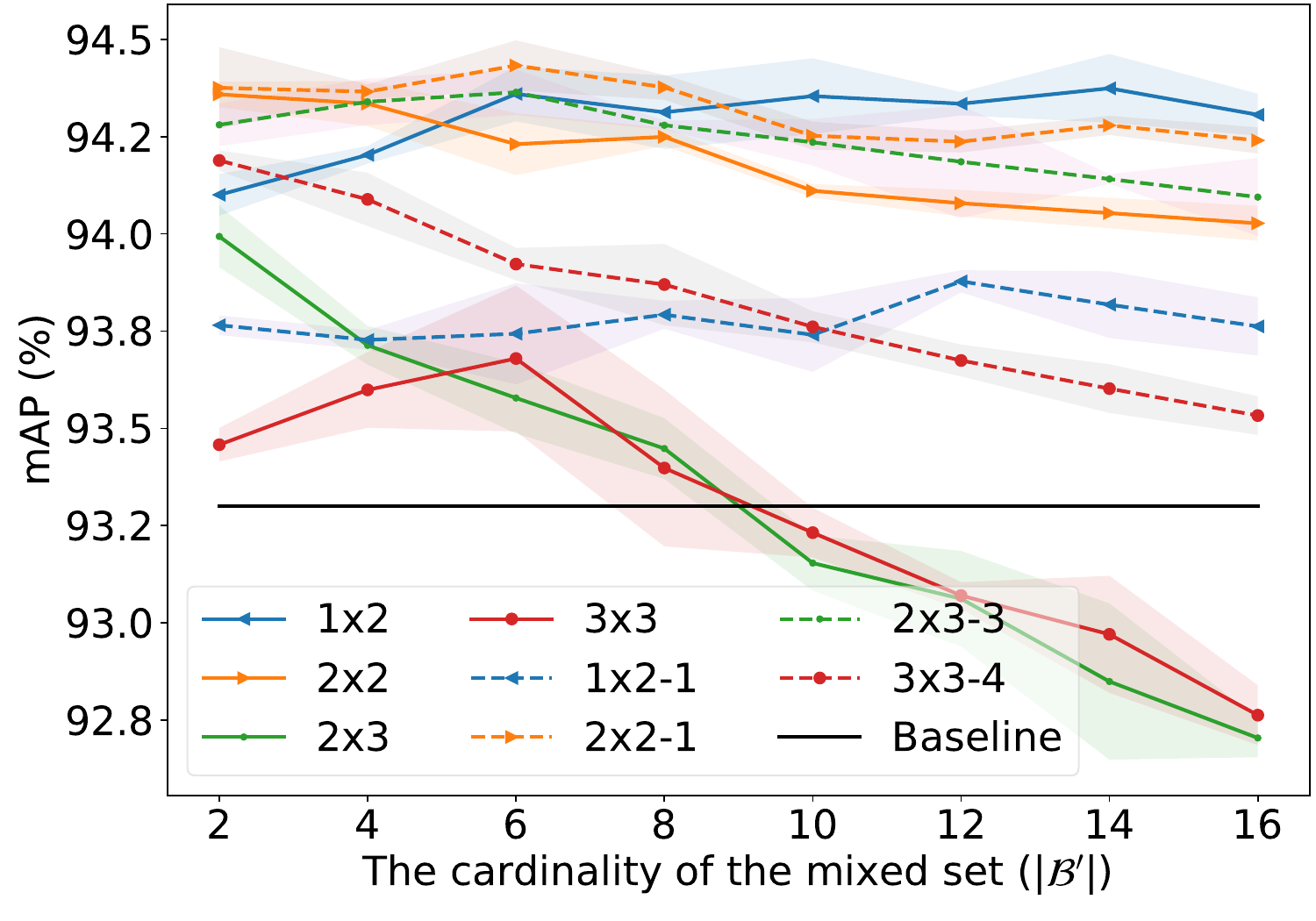}

   \caption{Comparisons of various grid settings on mAP metric. The mean in a solid or dash line and standard deviation in shaded region are given. Each solid-dash pair in the same color is a grid setting without and with dropout.}      
   \label{fig4.2}
\end{figure}

\ \\
\indent
Here, we discuss the influence on performance with different grid settings and number of mixed images. The results on Pascal VOC 2007 trained with the regular batch size of 16 are illustrated in \fig~\ref{fig4.2}. To obtain steady results, we run experiments three times and calculate the mean and standard deviation for each grid setting.

\par
Taking a look at the overview of \fig~\ref{fig4.2}, we can find that the great majority of grids outperform the baseline. Our \model\ is insensitivity to its parameters and only needs a few mixed samples to achieve remarkable performance. Secondly, we turn slights to the grids without dropout. As the number of mixed images increases, mAP of most grids descends. Too many mixed images, especially with large grids (\ie, $2\times 3$ and $3\times 3$), harm the learning ability of the model, due to the large scale discrepancy~\cite{touvron2019fixing} between regular images and their downsampled versions in mixed images, and the over-complicated semantic. 
% mixed images containing. 
For the latter, as stated in \sec~\ref{sec3.1}, we introduce dropout to reduce the semantic complexity. Comparing the pair of each grid setting without and with dropout in \fig~\ref{fig4.2}, we can see that dropout consistently improves its plain setting, except the grid of $1\times 2$. Actually, the grid of $1\times 2-1$ is degraded to involve only cross-scale learning.
% without semantic blending. 
This indicates that semantic blending plays a key role in our \model. Finally, we can summarize a referenced cardinality $|\mathcal{B}^\prime|_{ref}$ for the mixed set, that is $|\mathcal{B}^\prime|_{ref}=\lfloor|\mathcal{B}|/(r\times c-d)\rfloor$, where $|\mathcal{B}|$ denotes the regular batch size and $r$, $c$, $d$ denote the rows, columns, the number of dropped sub-images in a grid setting of $r\times c-d$, respectively.  
% Around the referenced cardinality
% According to \fig~\ref{fig4.2}, we also suggest that $r+c$ less than 6 will result in satisfying performance.  ## -> ILLOGICAL

\section{Conclusion} % (fold)
In this paper, we introduce a simple but effective augmentation strategy, namely \model, for multi-label image classification. \model\ augments the sample space and batch scale simultaneously. In augmented sample space, mixed images with semantic blending contribute to alleviating co-occurred bias. In augmented batch, the splice of regular images and mixed images enables cross-scale training and consistency learning. Hence, we also offer a non-parametric, consistency learning-based extension (\modelExt) to present the flexible extensibility  of our \model. Extensive experiments on various tasks demonstrate the effectiveness of the proposed methods.

\bibliographystyle{IEEEtran}
\bibliography{Refs_brief}

\newpage

\vfill

\end{document}